\documentclass[journal]{IEEEtran}\IEEEoverridecommandlockouts 
\ifCLASSINFOpdf

\else

\fi

\usepackage{xcolor,graphicx}       
\usepackage{subfigure}
\usepackage{indentfirst}

\usepackage{algorithm}
\usepackage{algorithmic}        
\usepackage{multirow}
\usepackage{array}

\usepackage{comment}          
\usepackage{epsfig}            
\usepackage{balance}
\usepackage{float}
\usepackage{caption}            

\usepackage{amssymb}

\usepackage{amsmath} \usepackage{amssymb}

\usepackage{cite} 
\usepackage{url}

\usepackage{multirow}
\usepackage{booktabs}
\usepackage{longtable}
\usepackage{bigstrut,multirow, rotating, booktabs}

\usepackage[colorlinks,bookmarksopen,bookmarksnumbered,citecolor=black,urlcolor=black]{hyperref}

\renewcommand{\eqref}[1]{Equation (\ref{#1})}

\usepackage[T1]{fontenc}

\begin{document}
	
	\title{A Survey on Trustworthy Edge Intelligence: From Security and Reliability To Transparency and Sustainability}
	
	\author{Xiaojie Wang, Beibei Wang, Yu Wu, Zhaolong Ning, Song Guo,~\IEEEmembership{Fellow, IEEE}, and Fei Richard Yu,~\IEEEmembership{Fellow, IEEE}
		
		\thanks{X. Wang, B. Wang, Z. Ning (Corresponding author) are with School of Communications and Information Engineering, Chongqing University of Posts and Telecommunications, Chongqing 400065, China. E-mail: wangxj@cqupt.edu.cn, s220101143@stu.cqupt.edu.cn, ningzl@cqupt.edu.cn.
			
			Y. Wu is with the School of Cyber Security and Information Law, Chongqing University of Posts and Telecommunications, Chongqing 400065, China. E-mail: wuy@cqupt.edu.cn.
			
			S. Guo is with the Department of Computer Science and Engineering, The Hong Kong University of Science and Technology, Kowloon, Hong Kong, China. E-mail: songguo@cse.ust.hk.
			

			
			F. R. Yu is with the Department of Systems and Computer Engineering, Carleton University, Ottawa, ON K1S 5B6, Canada. E-mail: RichardYu@cunet.carleton.ca.
			
		}
	}

	\maketitle

	\begin{abstract}

		Edge Intelligence (EI) integrates Edge Computing (EC) and Artificial Intelligence (AI) to push the capabilities of AI to the network edge for real-time, efficient and secure intelligent decision-making and computation. However, EI faces various challenges due to resource constraints, heterogeneous network environments, and diverse service requirements of different applications, which together affect the trustworthiness of EI in the eyes of stakeholders. This survey comprehensively summarizes the characteristics, architecture, technologies, and solutions of trustworthy EI. Specifically, we first emphasize the need for trustworthy EI in the context of the trend toward large models. We then provide an initial definition of trustworthy EI, explore its key characteristics and give a multi-layered architecture for trustworthy EI. Then, we summarize several important issues that hinder the achievement of trustworthy EI. Subsequently, we present enabling technologies for trustworthy EI systems and provide an in-depth literature review of the state-of-the-art solutions for realizing the trustworthiness of EI. Finally, we discuss the corresponding research challenges and open issues.
		
	\end{abstract}
	
	
	\IEEEpeerreviewmaketitle
	\IEEEpeerreviewmaketitle
	
	\section{Introduction}
	
	\indent With the development of Fifth-Generation (5G) wireless communication technologies, billions of wireless devices such as smartphones, sensors, and wearables can connect to the Internet. By 2025, the number of Internet of Things (IoT) devices are expected to reach 25.2 billion~\cite{9858872}, and these devices will generate enormous data with more than 79 Zeta bytes per year~\cite{9866918}.  Driven by the IoT, big data, and powerful computing, Artificial Intelligence (AI) has made further breakthroughs in intelligent applications such as natural language processing~\cite{liu2023pre}, computer vision~\cite{zhang2023growsp}, and robotics~\cite{10160687}. Especially in 2023, generative AI large models such as ChatGPT, DALL-E2, and DreamFusion, lead a trend that has become the focus of global attention. Generative large models based on billions of parameters can generate high-quality content in seconds, such as advertising images, short videos, writing copy, and voiceover. Currently, large models are mainly deployed on public clouds, mainly for training, with relatively small user sizes. However, the popularization of generative AI large models and the expansion of user scale will lead to a rapid increase in the computational demand for inference, exceeding the computational load for training. 
	
	
	\indent In order to improve efficiency, reduce cost, and consider data security and privacy, Edge Computing (EC)~\cite{8016573} has emerged. Compared to cloud computing, EC deploys computational resources closer to users and data sources at the network edge, and thus enables low transmission latency and high communication efficiency. Edge Intelligence (EI) combines EC with AI techniques to fully leverage the advantages and potential of both. Enterprises, including Google, Microsoft, Intel and IBM, have developed pilot projects to demonstrate the benefits of EC in paving the last mile of AI~\cite{zhou2019edge}. For instance, MediaTek's APU 790 chip has a built-in hardware-level generative AI engine that supports high-speed and secure EI computation, and deeply adapts the transformer model for sub-acceleration, capable of generating images in less than one second. In addition, the hybrid precision quantization and memory compression techniques enable high-end smartphones to run Large Language Models (LLMs) with up to 33 billion parameters. While Google's release of PaLM2, a large model, realizes the full link of AI from the cloud to the network edge. Its lightweight version of Gecko runs on cell phones, significantly improving inference efficiency, reducing service costs, making the model applicable to a lot of application scenarios and users, and further promoting the development of EI deployment.
	
	\indent These efforts facilitate a wide range of AI applications, from real-time video analytics~\cite{9723006}, Virtual Reality (VR), and Augmented Reality (AR) to smart healthcare~\cite{10066875}, Autonomous Vehicles (AVs)~\cite{9706268} and Unmanned Aerial Vehicles (UAVs)~\cite{3604933}. EI encompasses the interconnection and collaboration of large-scale smart devices, sensors, and edge nodes so that data can be collected, processed, and exchanged across domains to support the digital intelligence transformation of various applications and businesses~\cite{9606720}. In this context, humans, objects and intelligences are interconnected to form a complex networked ecosystem.

	
	\begin{table*}[tbp]
		\caption{Comparisons of related surveys on trustworthiness and EI.}
		\label{routing}
		\centering
		\begin{tabular}{|p{17mm}|c|p{40mm}|p{70mm}|c|} 
			\hline

			\multirow{2}*{\parbox[c]{17mm}{\centering\textbf{Topics}}} &
			\multirow{2}*{\parbox[c]{5mm}{\centering\textbf{Ref.}}} &
			\multirow{2}*{\parbox[c]{40mm}{\centering\textbf{Focuses}}} &
			\multirow{2}*{\parbox[c]{70mm}{\centering \textbf{Contribution}}} &
			\multirow{2}*{\parbox[c]{15mm}{\centering \textbf{Network scenarios}}}
			
			\\
			&&&& \\

			\hline
			
			\multirow{8}*{\parbox[c]{17mm}{\centering Trustworthiness}}&
			
			{\centering \cite{kaur2022trustworthy}} &
			\parbox[c][13mm]{40mm} {\centering Requirements for AI trustworthiness}&
			{\parbox[c][13mm]{70mm}{ A survey on AI risk migration and system validation technologies to make AI trustworthy from the perspective of human involvement.}}&
			
			{\centering Not specific}  
			\\
			\cline{2-5}
			& {\centering \cite{liu2022trustworthy}} &
			\parbox[c][13mm]{40mm} {\centering Technologies and applications for AI trustworthiness} &
			{\parbox[c][13mm]{70mm}{A review of representative technologies and real-world application examples for trustworthy ML from a computational perspective.}}&
			
			{\centering Not specific}  
			\\
			
			\cline{2-5}
			&{\centering \cite{li2023trustworthy}} &
			\parbox[c][13mm]{40mm} {\centering Enhancing trustworthiness in AI product development process} &
			{\parbox[c][13mm]{70mm}{ A review of theoretical frameworks and systematic approaches to improve AI trustworthiness from the perspective of the entire life cycle of AI systems.}}&
			
			{\centering Not specific}  
			\\
			
			\hline
			
			\multirow{18}*{\parbox[c]{17mm}{\centering EI}}&
			{\centering ~\cite{9985008}}&
			{\parbox[c][13mm]{40mm}{\centering Efficient DL inference for edge devices}}&
			{\parbox[c][13mm]{70mm}{ A review of state-of-the-art tools and techniques for efficient edge inference, highlighting the challenges and benefits of deploying DL models on resource-constrained edge devices.}}&
			
			{\centering Edge networks}  
			  
			\\
			\cline{2-5}
			
			&{\centering ~\cite{murshed2021machine}}&
		
			{\parbox[c][13mm]{40mm}{\centering Compression techniques, hardware and software, and applications}} 	&
			{\parbox[c][13mm]{70mm}{A survey of challenges and solutions for deploying ML models on IoT devices with limited resources at the network edge.}}&
			
			{\centering Edge networks}   
			\\
			
			\cline{2-5}
			&{\centering ~\cite{9134426}}&
			{\parbox[c][13mm]{40mm}{\centering Algorithms for efficient distributed training}}&
			{\parbox[c][13mm]{70mm}{A summary of the challenges and techniques for communication-efficient EI, emphasizing collaboration between edge devices and servers to overcome communication overhead.}}&
			
			{\centering Edge networks}  
			  
			\\
			\cline{2-5}
			&{\centering ~\cite{9606720}}&
			{\parbox[c][13mm]{40mm}{\centering The mutual support of advanced wireless technology and EI}}&
			{\parbox[c][13mm]{70mm}{A summary of solutions for EI in 6G networks when faced with challenges such as latency, energy consumption, network congestion, and privacy.}}&
			
			{\centering 6G networks}  
			  
			\\
			\cline{2-5}
			
			&{\centering ~\cite{9072101}}&
			{\parbox[c][13mm]{40mm}{\centering Security}}&
			{\parbox[c][13mm]{70mm}{A summary of security threats and attack surfaces for IoT systems is presented and various ML and DL algorithms applied to IoT security are reviewed.}}&
			
			{\centering  IoT}  
		 
			\\
			\cline{2-5}
			
			&{\centering ~\cite{10044183}}&
			{\parbox[c][13mm]{40mm}{\centering Security and privacy}}&
			{\parbox[c][13mm]{70mm}{A survey of security and privacy issues related to network edge server deployment in 6G networks from the perspectives of EC, edge caching, and EI, respectively.}}&
			
			{\centering  6G networks}  
			  
			\\
			\hline
			{\parbox[c]{17mm}{\centering Trustworthy EI}}
			&\parbox[c]{15mm}{\centering This survey}&
			{\parbox[c][13mm]{40mm}{\centering Addressing security, reliability, transparency and sustainability of EI}}&
			{\parbox[c][13mm]{70mm}{A survey on definition, characteristics, architecture, technologies, solutions, challenges and, open issues for trustworthy EI from the perspective of the combination of AI and EC.}}&
			
			{\centering Edge networks}   
			\\
			\hline
			
		\end{tabular}
		\label{tab1:summary}%
		\vspace{-5mm}
	\end{table*}
	
	\indent However, with the computational load gradually migrate from the cloud to the network edge, issues of constrained bandwidth, storage and computational resources become prominent. This results in a series of challenges for EI services, related to security, privacy, content generation accuracy, Quality of Service (QoS), and energy consumption~\cite{10172151}. These challenges directly impact the trust and acceptance of  services by stakeholders, which range from end-users to technology developers, from government regulation to the general public. Therefore, research on trustworthy EI is of great practical significance to improve service quality, ensure system security, and promote sustainable development.
	
\begin{figure*}[htbp]
	\begin{center}
		\includegraphics[scale=0.40]{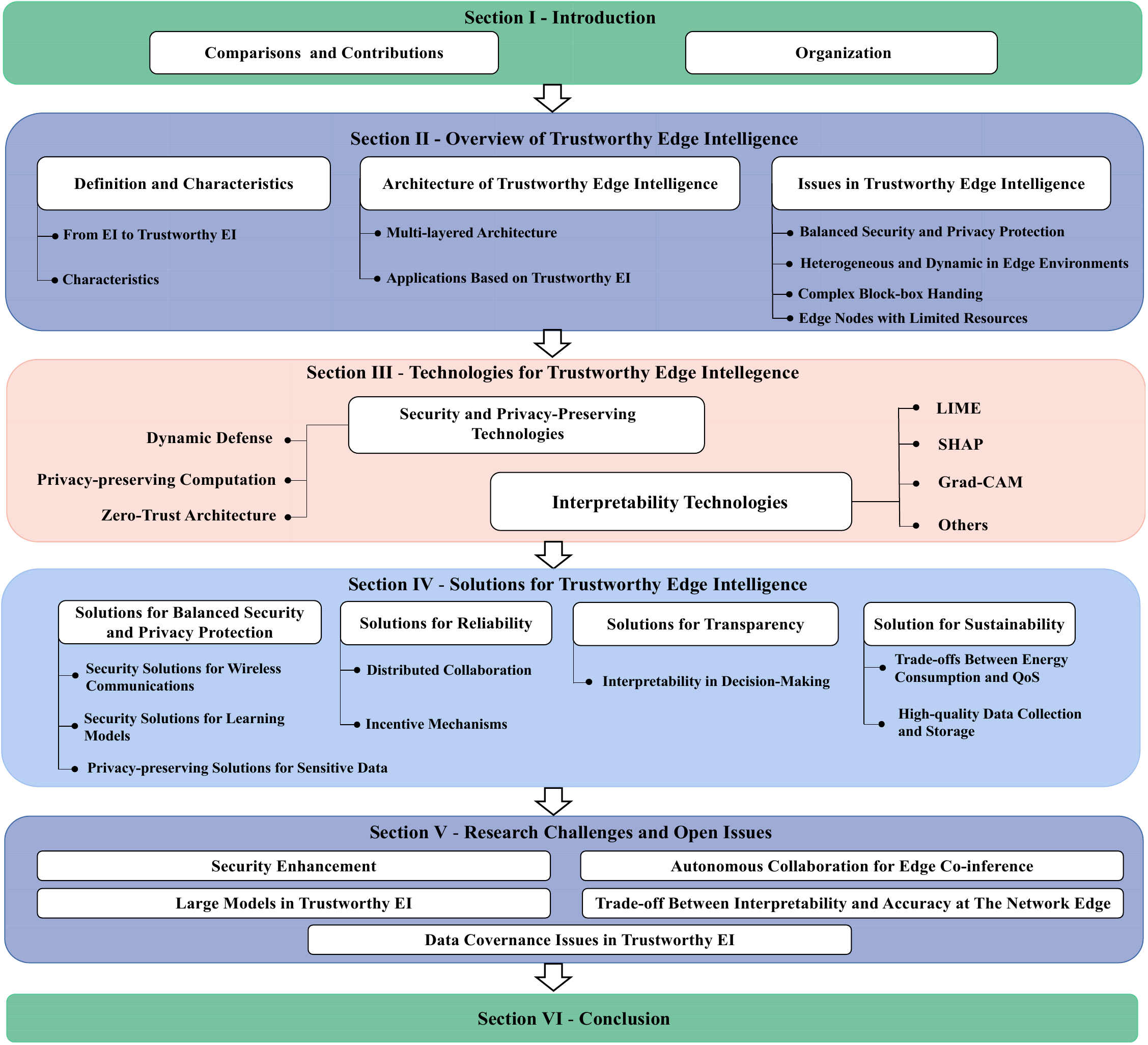}\\
		\caption{Structure of the survey.}\label{1:Structure of the survey}
	\end{center}
\end{figure*}

	\subsection{Comparisons and Contributions} \label{1}

	\indent Although several studies have been devoted to providing solutions for trustworthy EI in recent years, there has not been a comprehensive review and synthesis of techniques and solutions for trustworthy EI. 
	
	\indent As shown in Tab. \ref{tab1:summary}, some previous surveys focus on reviews of trustworthy AI~\cite{kaur2022trustworthy,liu2022trustworthy,li2023trustworthy}. Authors in~\cite{kaur2022trustworthy} provide various requirements for ensuring the trustworthiness of AI, along with corresponding methods, all from a human-centered perspective. Authors in~\cite{liu2022trustworthy} present a detailed review of representative techniques for trustworthy AI from a computational point of view and discuss their practical applications in real-world scenarios. Differently, authors in~\cite{li2023trustworthy} present a theoretical framework for important aspects of trustworthy AI from the perspective of the entire life cycle of an AI system. Meanwhile, they systematically introduce available methods for realizing trustworthy AI. However, the above surveys primarily concentrate on plausibility in scenarios where computational resources are centralized. In contrast, our survey is centered around scenarios at the network edge that demand specialized responses to unique challenges, such as limited resources and network instability.
	
	\indent Several surveys provide reviews of EI architecture, technologies and optimization algorithms. For example, the work in~\cite{9985008} presents a survey of software tools for hardware-algorithm co-design, along with an analysis of existing choices and trade-offs in hardware platforms. Authors in~\cite{murshed2021machine} delve into a range of applications related to Machine Learning (ML) and EC from an operational perspective. Furthermore, authors in~\cite{9134426} summarize communication efficient algorithms for distributed training of AI models. Authors in~\cite{9606720} focus on EI in 6G networks. 
	
	\indent Differently, researchers in~\cite{9072101} and~\cite{10044183} review security and privacy protection strategies at the network edge. To be specific, authors in~\cite{9072101} provide a review of the state-of-the-art ML and Deep Learning (DL) approaches from the perspective of IoT security. Authors in~\cite{10044183} explore security threats and countermeasures associated with EC, edge caching, and EI in the context of 6G network edges.
	
	\indent In summary, existing surveys focus on solving edge-efficient inference and training algorithms as well as security and privacy challenges. However, there is a lack of systematic survey that provide a comprehensive and professional discussion of the concepts, essential features, techniques, solutions, and challenges of trustworthy EI. \textit{\textbf{To the best of our knowledge, this survey is the first to provide a comprehensive summary of trustworthy EI from a combined EC and AI perspective.}} The contributions of this survey can be summarized as follows:
	\begin{itemize}
		\item We first provide a definition of trustworthy EI, its essential characteristics, and establish a three-layer architecture to support the concept. We then summarize some key issues with respect to the vulnerability of edge networks and AI models to highlight the difficulties faced in achieving trustworthiness of EI.
		
		\item We introduce enabling techniques for trustworthy EI in terms of security and interpretability, respectively.
		
		\item We provide a comprehensive and in-depth investigation of recent studies on trustworthy EI based on its issues and requirements. In addition, lessons learned for each kind of approaches are also provided.
		
		\item Finally, we provide a detailed discussion about our vision for the future development of trustworthy EI, and identify a number of open challenges that may give rise to promising research directions.
	\end{itemize}
	
	\subsection{Organization}
	
	\indent As shown in Fig. \ref{1:Structure of the survey}, the survey is organized as follows. Section \uppercase\expandafter{\romannumeral2} examines the concept, characteristics, architecture, and issues of trustworthy EI. We present technologies used to achieve trustworthiness in Section \uppercase\expandafter{\romannumeral3}, and discuss solutions to realize trustworthy EI in Section \uppercase\expandafter{\romannumeral4}. Challenges and open issues for trustworthy EI are provided in Section \uppercase\expandafter{\romannumeral5}, and the survey is concluded in Section \uppercase\expandafter{\romannumeral6}. 
	
	
	\section{Overview of Trustworthy Edge Intelligence}\label{Trustworthy Edge Machine Learning}
	
	\indent This section provides a comprehensive overview of trustworthy EI, covering its definition, characteristic, architecture, cutting-edge application scenarios and issues. 
	
	
	\subsection{Definition and Characteristics}
	
	\indent In this following, we begin by examining the problems faced by EI, clarifying the urgency of trustworthy research. Subsequently, we explicitly define the concept of trustworthy EI and the fundamental characteristics it possesses.

	\subsubsection{From EI to Trustworthy EI}
	
	
	\indent In the actual deployment and application of EI, users and organizations remain skeptical and cautious as it introduces several concerns and risks: \romannumeral1) Security and privacy concern: In open edge environments, communication vulnerabilities may lead to sensitive data leakage, while AI algorithms are susceptible to poisoning and adversarial attacks, especially large models that are pre-trained and fine-tuned at the network edge, increasing the risk of poisoning and adversarial attacks. As a result, users may lose confidence in EI due to concerns about the unreliability of models in mission-critical situations, such as parking and acceleration decisions for AVs;
	\romannumeral2) Imbalanced performance: Limited computational resources, restricted storage capacity, and limited energy supply of edge devices lead to imbalance in the performance of EI services, especially in real-time applications. Because reducing the model complexity to adapt to resource constraints may bring about degradation in model accuracy and may even increase the risk of security and legal liabilities; \romannumeral3) Un-interpretable model: Black-box models, such as LLMs usually set a large number of parameters to boost inference performance, making it difficult for users to understand how the model works, its limitations, and potential flaws. Even open source tools like LLaMA provide only limited interpretability. Interpreting and publicizing training data becomes very complicated, especially for proprietary models such as ChatGPT and Claude, whose architectures and training data are not yet publicly available \cite{165498}. 
	
	\indent These issues restrict the widespread application of EI and introduce a trust deficit for the digital network ecosystem. Therefore, it is important to investigate trustworthy EI to alleviate the trust concerns. Before defining trustworthy EI, we draw on the insights from~\cite{kumar2020evolution} to briefly discuss the concepts, distinctions, and connections between trust and trustworthiness. Although their precise definitions depend on specific application contexts, there is a general and broad consensus that trust is a psychologically and emotionally dimensional concept involving an individual's or organization's confidence in and expectations of another person, organization, or system. Trust is formed when an individual or organization feels reliable about a party's behavior, commitment, or competence. In contrast, trustworthiness focuses on the objective level and relies on the reliability, stability, and security of a system, service, or individual in its design, implementation, and operation. A trustworthy entity should be able to fulfill its commitments, protect the interests of others, and demonstrate reliability and accountability in all aspects. Thus, judgments of trust reflect beliefs about the trustworthiness of another party. Ensuring that the entity has a high degree of trustworthiness is the basis for building trust among users and other stakeholders.
	
	\indent In the context of EI, this entity represents systems that integrate EC, IoT devices, edge servers, and AI models deployed at the network edge. It aims to enable real-time intelligent decision-making and computation locally or in close proximity to the data source. Therefore, trustworthy EI can be defined by \textbf{\textit{"A EI system employs a variety of technologies and strategies to ensure the system to achieve the predefined goals during the design phase while minimizing potential risks and hazards"}}.

	\begin{figure*}[htbp]
		\begin{center}
			\includegraphics[scale=0.48]{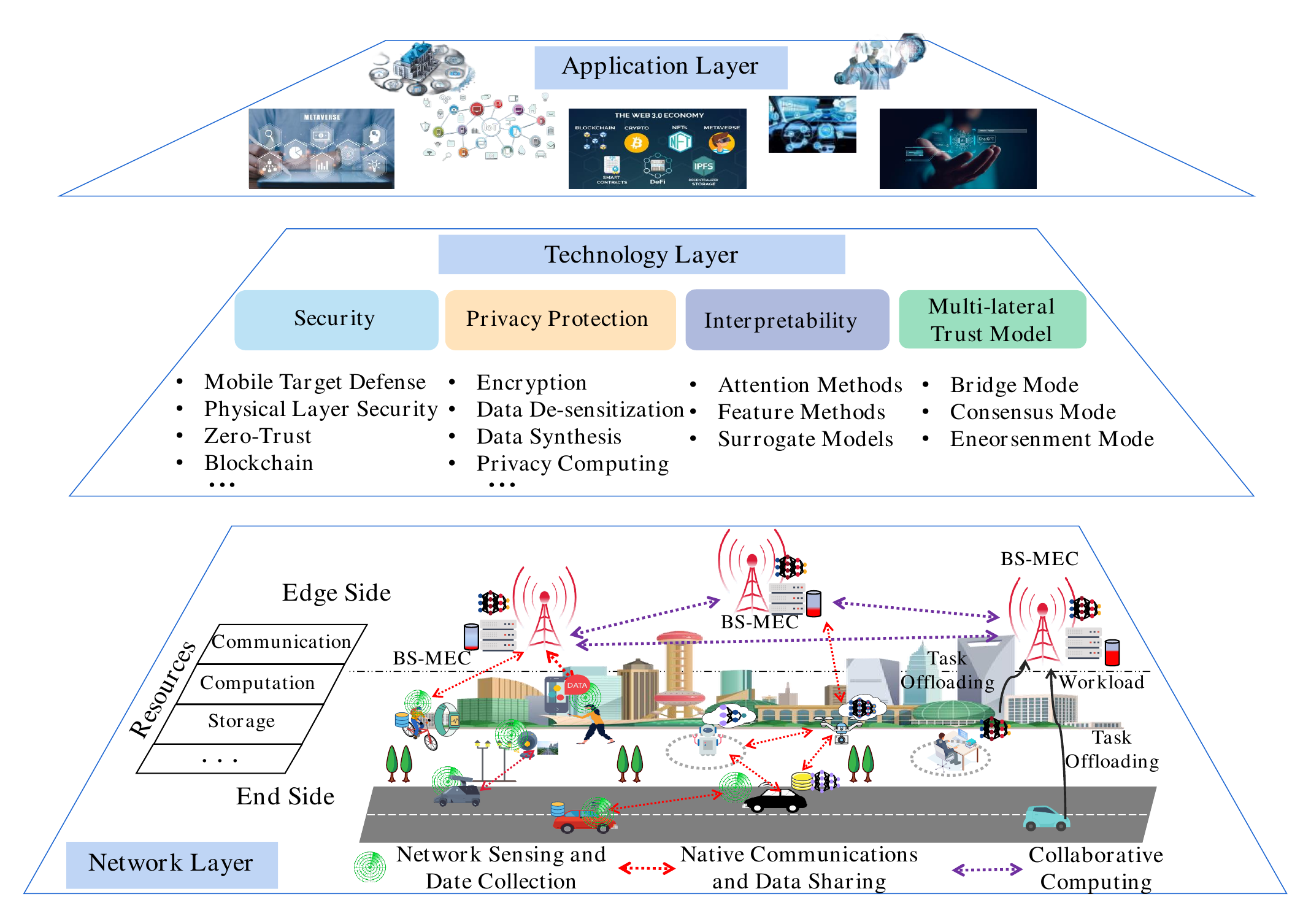}\\
			\caption{Architecture of trustworthy EI: The network layer integrates communication, computation, perception, and intelligence, delivering the application layer high-performance, interpretable decision-making, resilient and sustainable solutions. Meanwhile, the technology layer establishes a trustworthy environment, offering robust technical support across the network and application layers, fostering seamless collaboration for comprehensive trustworthy EI.}\label{2:ss}
			\vspace{-5mm}
		\end{center}
	\end{figure*}

	\subsubsection{Characteristics}
	
	\indent While considering the unique features of EI, such as limited resources, real-time requirements, and distributed deployment, and simultaneously conducting a comprehensive evaluation of the overall performance of EI systems, the characteristics of trustworthy EI can be briefly summarized as follows: Endogenous security, reliability, transparency, and sustainability. These characteristics not only address current issues faced in EI systems but also effectively respond to potential new challenges in the future, enabling the system to possess enhanced adaptability and garner trust from stakeholders. We elaborate these characteristics in the following content:
	
	\begin{itemize}
		\item \textbf{Endogenous Security:} In the field of EI, traditional security approaches face the dilemma of coping with the challenges of real-time, distributed, and limited resources. Endogenous security emphasizes that security mechanisms should have the ability to proactively adapt to dynamic changes at the network edge with lightweight deployment. By introducing technologies and policies such as Physical Layer Security (PLS), zero trust, and anomaly detection, trustworthy EI systems are able to provide cost-effective, efficient, and reliable security in areas such as wireless communications, distributed learning, and data sharing. The endogenous security breaks through the limitations of traditional security methods, emphasizes the inherent initiative and intelligence of the system, and brings a new perspective to the security of trustworthy EI systems.
		
		
		\item \textbf{Reliability:} It typically involves the system's fault-tolerance capability and performance consistency. Different from traditional networks, trustworthy EI systems place special emphasis on real-time requirements, resource constraints, and distributed computing. Firstly, many EI-driven applications such as AVs, AR, and VR demand high real-time performance and low latency~\cite{xiao2022perception}. The trustworthy EI must ensure real-time inference for quick decision-making within extremely short timeframes. This not only depends on communication methods and algorithm complexity but also relies on efficient resource scheduling on edge devices. In contrast, core networks usually operate in large data centers with relatively abundant computing and storage resources, may prefer the reliability of communication links. Lastly, reliability needs to account for the collaborative execution of tasks across different edge nodes to ensure the overall stability of the system.
			
			

		\item \textbf{Transparency:} In trustworthy EI system, transparency encompasses not only the interpretability of decision-making process but also clear data presentation. First, it is crucial to ensure that the decision-making process is interpretable. Given the limited resources of edge devices, interpretable techniques and strategies must strike a balance, ensuring that explanatory information is both comprehensive and adheres to computational constraints~\cite{9349455}. Second, transparency in data usage allows users to understand key information about the data source, quality, and processing within the system. This transparency is essential for establishing user trust, ensuring privacy compliance, and enhancing system credibility. 
		
		

		\item \textbf{Sustainability:} It reflects the long-term health and stability of the trustworthy EI system. On the one hand, edge devices rely on limited energy sources such as batteries, and the use of energy-efficient algorithms can extend device lifespan and reduce environmental stress~\cite{zhu2021green}. In addition, the distributed infrastructure of EI systems includes outdoor environments where adverse environmental impacts can be reduced through low-power hardware and green computing practices. On the other hand, EI utilizes edge devices to continuously sense and collect data, emphasizing the importance of high-quality data resources to ensure model accuracy and reliability~\cite{9714882}. In conclusion, sustainability not only promotes environmental friendliness, but also enables trustworthy EI systems to provide long-lasting and reliable services while maintaining peak performance.

	\end{itemize}

	\indent In summary, the above characteristics constitute the trustworthiness of EI, fostering trust between the EI system and stakeholders. They are usually interrelated rather than independent. In resource-limited edge networks, security measures should be lightweight. Reliability allows the system to balance performance and safety, while security enhances resistance against external threats, improving overall reliability and transparency. In addition, transparency in decision-making empowers users and developers to understand how the model works and facilitates prompt adjustments in case of errors. A secure, reliable, and transparent system is more likely to sustain in a dynamic environment, showcasing mutual reinforcements and trade-offs among these characteristics for trustworthy EI.

	\subsection{Architecture of Trustworthy Edge Intelligence} \label{privacy}
	
	\indent In this subsection, a multi-layered architecture for scalable and trustworthy EI systems is presented. Following, we detail the components of this architecture and the supported preamble application services.

	\subsubsection{Multi-layered Architecture}\label{ooo}
	
	\indent As shown in Fig. \ref{2:ss}, the architecture is divided into the network layer, the technology layer and the application layer.
	
	\indent The network layer, as a basic and key level, covers edge and IoT devices. Its main task lies in realizing cooperative communication among devices and accomplishing data collection, transmission, processing and analysis at the network edge, so as to build the foundation of EI service~\cite{9866918}. Among them, end devices with embedded sensing, communication and limited processing capabilities collect data in the environment, establish connections, exchange data and share models with other devices at the network edge. Edge nodes include fog nodes, gateways, and EC servers. Compared to end-side devices, edge nodes have more powerful storage and computation capabilities and can provide high-quality network and processing services with low latency. The edge-side ones are mainly responsible for end-side and edge-side resource fusion, coordinating heterogeneous resources to achieve fast and flexible service deployment and low-latency task processing.

\begin{figure}[tbp]
		\begin{center}
			\includegraphics[scale=0.5]{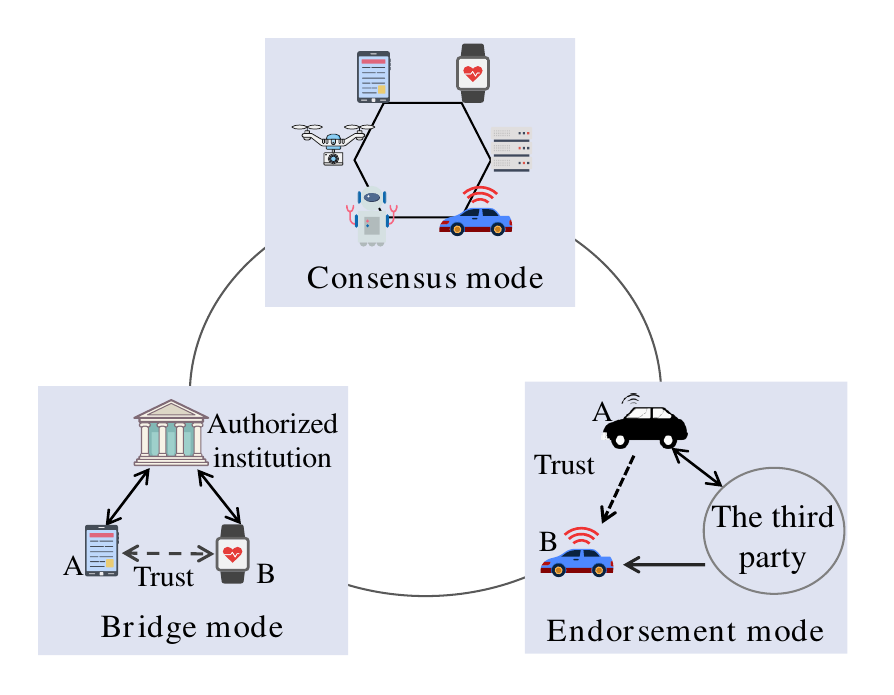}\\
			\caption{The multi-lateral trust model.}\label{3:sss}
			\vspace{-10mm}
		\end{center}
	\end{figure}

	\indent The technology layer plays an integral role in trustworthy EI, providing critical trustworthy technology support to the network layer. By ensuring high-performance services, sustainable and interpretable intelligent decision-making, and a reliable and resilient network, the technology layer safeguards the trustworthiness and security of the entire system. In particular, the multi-lateral trust model plays a pivotal role in establishing a robust foundation of trust for communication, collaboration, and information sharing within the network layer. It achieves this by creating and managing a framework for trust relationships among devices. The model encompasses three key modes: bridge, endorsement, and consensus, as shown in Fig. \ref{3:sss}.
	
	\indent In the "bridge" model, a central authorization authority conducts peer-to-peer authentication between entities A and B, facilitating the direct establishment of a trust relationship. The "endorsement" mode relies on a third-party organization to assess the trustworthiness of an entity, subsequently transmitting the evaluation results of entity A to entity B to establish trust. The "consensus" mode is considered as the most crucial one, employing distributed transactions among entities. This model offers a flexible and scalable solution for building trust relationships in a distributed manner.
	
	\indent At last, the application layer makes full use of the trustworthy technologies and infrastructure provided by the network layer to realize high-quality, well-accepted and trustworthy application services. In the following subsection,  we elaborate on several applications in detail, demonstrating the benefits of trustworthy EI in these emerging scenarios.

	\subsubsection{Applications based on Trustworthy EI}
	
	
	
	\indent Trustworthy EI is at the forefront of innovative applications in multiple cutting-edge fields. From mobile Artificial Intelligence-Generated Content (AIGC) to Web 3.0 and onward to Metaverse, these domains are continuously expanding the boundaries of our digitized lives. In the following, we delve into these trendsetting application areas and analyze the key role of trustworthy EI in ensuring the secure, efficient, and reliable deployment of mobile AIGC, Web 3.0, and the Metaverse.

	\begin{itemize}
	
		\item \textbf{Mobile AIGC:} AIGC has become a novel approach for processing and manipulating data, supporting multi-modal input and output to automatically generate creative and personalized content. The mobile AIGC is the deployment of AIGC services in mobile edge networks to enhance the user experience of mobile applications and generate innovative solutions~\cite{xu2023unleashing}. The trustworthy EI architecture plays a crucial role to ensure efficient, accurate, and secure mobile AIGC services. Specifically, in the trustworthy EI, communication and distributed computational resources at the network edge and the end side are integrated and allocated in real time, enabling mobile AIGC applications to quickly adapt to user demands, maximize the utilization ratio of limited resources, and maintain accurate content generation. Lightweight security and privacy protection mechanisms ensure that AI model training is safe while minimizing the use of computational resources, thus contributing to improving efficiency, security, and accuracy of mobile AIGC services. Therefore, the reliability and security characteristics of trustworthy EI strongly support the secure and sustainable deployment of mobile AIGC.
		
		
		\item \textbf{Web 3.0:} It is considered that the next generation of the Internet allows users to read, write, and own content~\cite{10061682}. Current Web 3.0 services primarily focus on supporting blockchain applications, including Decentralized Identifiers (DIDs), digital asset management, Decentralized Applications (DApps), cryptocurrency-based Decentralized Finance services (DeFi), and Metaverse. Trustworthy EI plays a crucial role in realizing the decentralized, secure, and user-controlled Web 3.0. First, it allows data processing on edge devices, providing Web 3.0 users with a secure means to own and control their digital identity and data. Second, decentralized applications in Web 3.0 may involve edge devices such as IoT devices and sensors. The trustworthy EI enables these devices to efficiently and reliably collaborate in executing smart contracts and processing local data, thereby reducing network latency. In summary, trustworthy EI provides essential technical support for the implementation of Web 3.0, contributing to the creation of a decentralized, user-friendly, and secure digital ecosystem.
		
		\item \textbf{Metaverse:} It is considered as the evolutionary paradigm of Web 3.0. This concept integrates a plethora of existing technologies, including 5G, Multi-access Edge Computing (MEC), AI, VR, blockchain, digital currency, IoT and human-computer interaction~\cite{9944862}. Trustworthy EI significantly enhances the computational efficiency, accuracy, scalability, privacy, and security of AI-driven virtual services in the Metaverse, including AR/VR recommendation and cognitive virtual identities. Specifically, trustworthy EI supports distributed collaborative computational and real-time resource allocation, markedly improving the processing efficiency of tasks such as high-dimensional data processing, 3D virtual world rendering, and avatar computation in the Metaverse. Edge learning with secure aggregation mechanisms facilitates local training, effectively reducing communication costs associated with AR content delivery, such as 3D objects or high-resolution video streams, and ensuring the security of model training and privacy of sensitive data. Furthermore, trustworthy EI with transparency helps users to understand why the system makes a particular recommendation or behavior, thus increasing their trust and acceptance of the virtual service. In summary, the trustworthy EI is crucial for ensuring QoS, user privacy, security, and stable operation of Metaverse applications.
	\end{itemize}

	\subsection{Issues in Trustworthy Edge Intelligence} 
	
	\indent Although trustworthy EI can provide many benefits, in the face of resource-constrained edge networks and complex AI models, in order to realize the trustworthiness of EI, the following issues need to be considered.
	

	\subsubsection{Balanced Security and Privacy Protection} 
	
	\indent The escalating demand from users for real-time and seamless services and requests necessitates that trustworthy EI systems strike a balance between security and privacy protection capabilities and user experience~\cite{10044183}. Excessively strict security policies may result in resource wastage, reduced system flexibility, and even negative impacts on user privacy. By appropriately configuring network architecture and employing intelligent security mechanisms, the system can effectively resist potential security risks while maintaining a high level of network and service quality. Hence, a flexible and balanced security strategy becomes particularly crucial in trustworthy EI environments. We discuss solutions to solve this issue in subsection \uppercase\expandafter{\romannumeral4}-A.
	
	\subsubsection{Heterogeneity and Dynamics in Edge Environments} 
	
	\indent The differences in computing capability, storage capacities, network bandwidth, sensing abilities, and domain knowledge of each node make the management and scheduling of resources extremely complex. Therefore, it is necessary to further consider how to coordinate and fully utilize different resources and knowledge of nodes to improve the performance and effectiveness of the whole system. In addition, the resource state of nodes changes over time, including device connection and battery power, further increases the difficulty of resource management. This requires the system to be able to flexibly respond to the dynamic changes of nodes to ensure the normal operation at different time and conditions. We thoroughly examine solutions to this issue in subsection \uppercase\expandafter{\romannumeral4}-B.
	

	\subsubsection{Complex Black-box Handling} 
	
	\indent According to the previous discussion, to enhance the transparency of black-box models, interpretable methods need to be provided. However, increasing interpretability often introduce additional computational and storage costs, which can be a challenge for the limited resources of edge devices. Interpretability methods for complex black-box models may entail sacrificing real-time performance. Therefore, it is important to balance the interpretability requirements and real-time performance. We delve into detailed solutions for this issue and provide a summary of relevant literature in subsection \uppercase\expandafter{\romannumeral4}-C.
	

	\subsubsection{Edge Nodes with Limited Resources} 
	
	\indent Edge devices are often constrained by limited computation and storage resources, and executing complex AI inference and training tasks may consume a significant amount of these resources, thereby reducing the lifetime of these devices. Specific EI applications, such as AVs and smart factories, have extremely high requirements for latency, which may result in high power consumption, adversely affecting energy efficiency. In addition, certain intensive AI tasks can only be performed on cloud servers, which causes high communication costs. Thus, leveraging ubiquitous network resources while reducing the communication overhead becomes a major bottleneck for AI applications at the network edge. We discuss some possible solutions for this issue in subsection \uppercase\expandafter{\romannumeral4}-D.
	
	
	\section{Technologies for Trustworthy Edge Intelligence}\label{Technologies for Trustworthy Edge Machine Learning}
	
	
	\indent In this section, we focus on technologies that play a crucial role in ensuring the trustworthiness of EI. The utilization of these technologies serves as a means to address existing challenges and enhance system trustworthiness. 
	
	\subsection{Security and Privacy-Preserving Technologies}
	
	
	\indent The devices involved in EI are usually deployed close to the user side to realize real-time data processing and decision making. Both the security of computation and the risk of sensitive data leakage may reduce user trust. Therefore, security and privacy-preserving technologies can be utilized to minimize the susceptibility of EI systems to attacks and prevent data leakage, thereby enhancing user confidence in the system. Next, we provide a detailed description of security and privacy-preserving techniques, respectively.
	
	\subsubsection{Dynamic Defense}


	\indent With the automation and intelligence of adversarial techniques continually advancing, dynamic defense technologies gain widespread attention among researchers. The idea is to dynamically alter and disguise characteristics of target systems, thereby increasing attack costs, enhancing the system resilience, and bolstering defense capabilities. In the following, we introduce these related defense technologies in detail, which include Cyberspace Mimic Defense (CMD), Mobile Target Defense (MTD), and Cyber Deception (CD).
	
	\begin{itemize}
		
		\item \textit{Cyberspace Mimic Defense:} The CMD aims to prevent attackers from forming effective attacks by means of conditional evasion, so that the inevitable endogenous security problems do not become security threats to the system. The core idea is to organize multiple redundant and heterogeneous methods to jointly process external requests, and achieve dynamic scheduling through negative feedback to compensate for security flaws in the cyberspace.
		
		\item \textit{Mobile Target Defense:} The MTD is an active defense mechanism that prevents network attacks by continuously and dynamically changing attack surface~\cite{8949517}. Its objective is to create uncertainty for attackers and shift the asymmetry between attackers and defenders. 
		
		\item \textit{Cyber Deception:} Compared to MTD, CD employ more aggressive strategies, intentionally providing false information (such as baits and honeypots) to mislead attackers~\cite{9610063}.

	\end{itemize}
	
	\subsubsection{Privacy-preserving Computation} 
	
	\indent Ensuring data privacy usually requires the use of different techniques and methods, such as Differential Privacy (DP), Homomorphic Encryption (HE), Secure Multi-party Computation (SMC), and Trusted Execution Environment (TEE). In the following, we provide a brief introduction to them.
	
	\begin{itemize}
		
		\item \textit{Differential Privacy:} DP aims to provide statistical guarantees for individual data while minimizing the disclosure of individual privacy~\cite{9069945}. The main idea is to add noise, such as Laplace noise, to the original query results (numerical or discrete values). The added noise prevents the inference of significant information about individuals from query results, preserving personal privacy. 
		
		\item \textit{Homomorphic Encryption:} HE is a class of encryption mechanisms that support processing and computation of ciphertexts~\cite{acar2018survey}. In actual training and inference process, HE techniques can ensure the security of model parameters and raw data, thus developing trustworthy EI models~\cite{10250861}. However, existing HE solutions need to address the problem of high computational overhead, especially in resource-constrained edge environments.
		
		
		
		\item \textit{Secure Multi-party Computation:} SMC enables collaborative computation on a combined dataset without compromising the data privacy of individual parties~\cite{zhang2022lsfl}. By implementing SMC at the edge nodes, trustworthy collaborative computation can be achieved, and data availability without data visibility can be guaranteed.
		
		\item \textit{Trusted Execution Environment:} It is an isolated processing environment designed to provide computing and storage capabilities with security and integrity guarantees. The basic idea is to allocate segregated hardware memory for sensitive data, ensuring secure transmission, storage, and processing. Therefore, deploying TEE can ensure the confidentiality of both EI models and input data~\cite{zhao2021sear}.
	
	\end{itemize}
	
	\subsubsection{Zero-Trust Architecture}
	
	\indent The fundamental principle of zero-trust is to distrust any access, whether inside or outside the network, and to advocate for continuous authentication and dynamic authorization to ensure global defense. Zero-trust network access relies on micro-segmentation and network isolation, eliminating the need for a virtual private network by granting access to the network only after thorough verification and authentication. 
	
	
	\indent The increased connectivity at the network edge inevitably expands attack surface, allowing attackers to access systems through multiple entry points. Additionally, not all devices can receive security updates promptly, potentially enabling attackers to exploit multiple vulnerabilities to gain access to the network. Therefore, applying the zero-trust principle to edge networks forms a new defense boundary, capable of integrating security and networking anywhere~\cite{10000958}. Zero-trust edge, based on continuous verification of user identity and context, provides explicit access to applications. 
	
		
		
		
		
	\subsubsection{Others}

	
	\indent \textit{Physical Layer Security (PLS)} is used to enhance the security of wireless communication systems. Different from traditional encryption methods, PLS relies on physical properties of communication channels rather than algorithms and keys. Common PLS techniques include: artificial noise~\cite{wang2023latency}, cooperative jamming~\cite{9980450}, and beamforming~\cite{9206080}. Specifically, artificial noise is used to mask original data from eavesdroppers by intentionally adding noise to the transmitted signal in the communication channel. In cooperative jamming, multiple nodes work together to protect communication privacy by interfering with potential eavesdroppers. Beamforming technology concentrates signal energy in a specific direction while reducing signal strength in other directions, making it difficult for eavesdroppers to intercept.
	
	\indent \textit{Anomaly detection} is used to automatically identify abnormal data. During the process of edge training, data updates from various training nodes can be analyzed to detect malicious updates based on differences between pairs of remote updates~\cite{shi2021federated}. This process enables the acquisition of a robust global model.
	
	
		\indent \textit{Blockchain} is a cryptographic, decentralized, and user-transparent technology that provides secure transactions and computing at the network edge~\cite{9808390}. Compared to the above techniques, blockchain has unique technical features such as consensus protocols and distributed ledgers. These features enable blockchain to effectively regulate security risks in EI systems. First, the consensus mechanism ensures the establishment of trust among devices for model training. Second, the tamper-proof distributed ledger keeps the recording of authentic and reliable information, promoting a transparent process~\cite{9761745}. Last, it rewards participating nodes based on their contributions, which incentivizes selfish nodes to provide their local resources~\cite{wang2022infedge}.
		
		%

	
	\subsection{Interpretability Technologies}\label{Interpretable Technologies}
	
	\indent Local Interpretable Model-agnostic Explanations (LIME) and SHapley Additive exPlanations (SHAP) are applicable to various models, while Gradient-Weighted Class Activation Mapping (Grad-CAM) specifically targets Convolutional Neural Network (CNN) models. In the following, we offer a brief overview of these technologies.

	\subsubsection{LIME} It is a model-agnostic interpretability method, and focuses on interpreting individual instances rather than the entire model to provide local explanations for specific test inputs~\cite{ribeiro2016should}. It introduces random perturbations to the input instances of a black-box model and trains an interpretable surrogate model, such as decision trees and linear models. The weights of the surrogate model can directly reflect the significance of features and their impacts~\cite{9464703}. To ensure interpretability and local fidelity, LIME tries to minimize the discrepancy between the surrogate model and the black-box model at the instance point.

	\subsubsection{SHAP} It can be regarded as a unified approach that combines LIME and shapley values~\cite{lundberg2017unified}. The shapley value of a single feature is the weighted average of the marginal contribution of that feature to a subset of all feature combinations. SHAP is the basis for fairly distributing contributions of each feature to the model and has three desirable properties (i.e., local accuracy, missingness, and consistency)~\cite{lundberg2020local}. Generally, SHAP is applied as an interpretability method based on feature correlation in edge scenarios such as smart healthcare~\cite{9478224} and intrusion detection~\cite{9830113}, not only to interpret final decisions, but also to support industry experts to quickly optimize and evaluate the correctness of their judgments.
	
	

	\subsubsection{Grad-CAM} It is a visual local interpretability method designed specifically for CNN models~\cite{selvaraju2017grad}. It calculates the gradient of the target class with respect to the last convolutional feature map of the CNN. This gradient information helps to identify image regions that have the biggest contribution to model predictions. By back-propagating gradients and multiplying them with the feature map, the importance weight is assigned to each pixel. These weights are then used to generate a heat map, which highlights the significant regions in the image. 
	
	
	\indent Grad-CAM supports a wide range of CNN models and does not require further changes to the model architecture. Since the size of feature maps is usually much smaller than the input image, heat maps produced by Grad-CAM may not provide precise localization. Authors in~\cite{selvaraju2017grad} combine a fine-grained visualization method of guided     back-propagation with Grad-CAM  to produce high-resolution activation maps. The guided Grad-CAM tries to provide intuitive and effective interpretation in clinical medical image analysis, helping physicians to determine the location, type, and severity of lesions, and facilitating the advancement and application in the field of medical imaging analysis~\cite{9435063}.
	
	\subsubsection{Others}
	
	\indent \textit{Attention mechanism} is a widely used technique that mimics features in the human visual and perceptual system to process and interpret input data. Visual attention is a method for visualizing the attention weights of a model, usually for text and image data.

	
	\indent \textit{Rule-based} interpretability techniques explain the behavior of a model by defining a set of rules. These rules can be created manually or generated by automatic learning techniques. Representative techniques include decision trees, rule sets, and expert systems.

	\section{Solutions for Trustworthy Edge Intelligence}\label{Solutions for Trustworthy Edge Machine Learning}
	
	\begin{figure*}[htbp]
		\begin{center}
			\includegraphics[scale = 0.5]{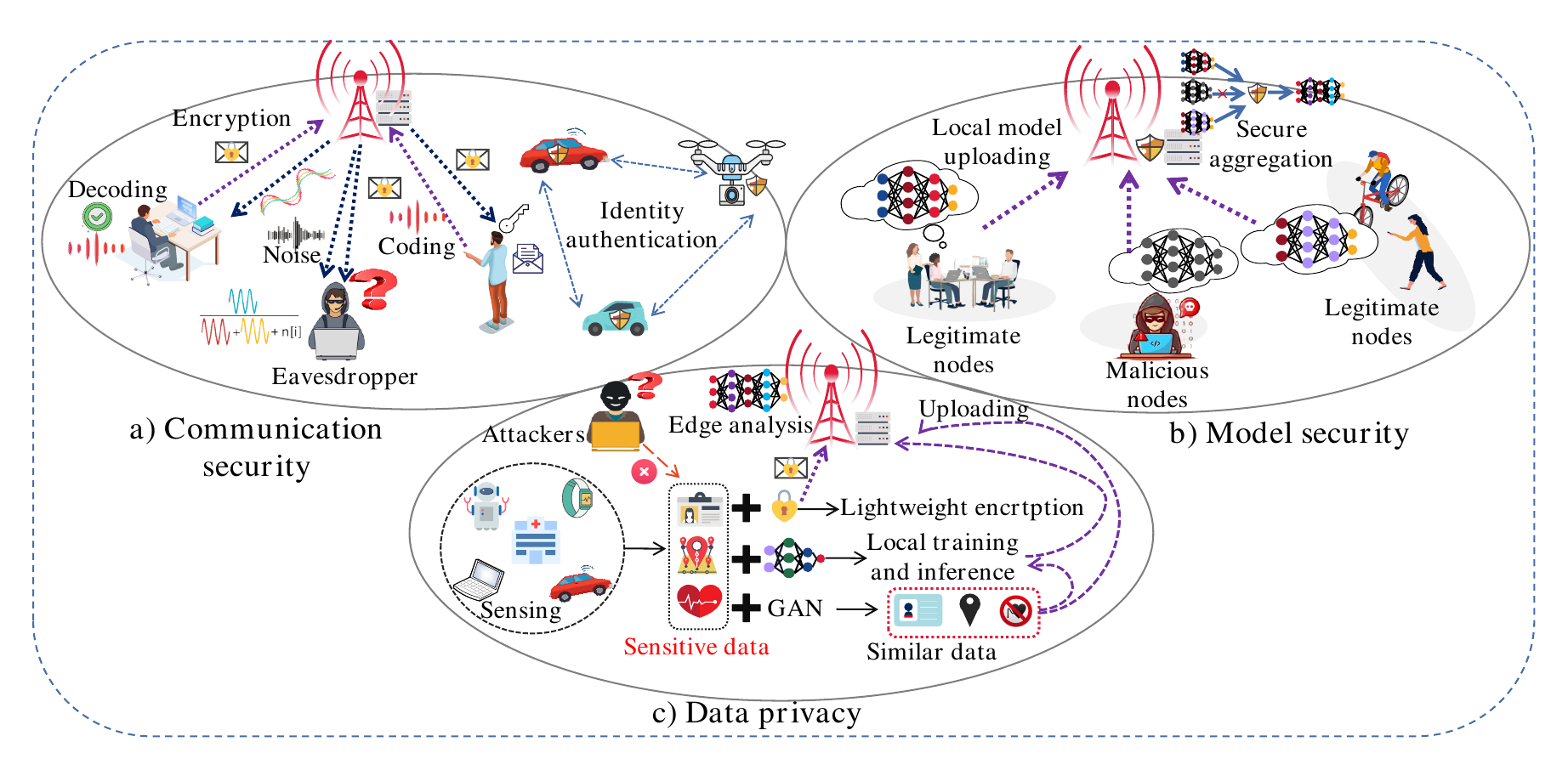}\\
			\caption{Illustrative solutions of balanced security and privacy protection: a) Security solutions for wireless communications ensure availability of the edge network and security of data transmission; b) Security solutions for learning models address threats such as model poisoning and adversary attacks, ensuring trustworthiness and robustness of learning models; and c) Privacy-preserving solutions focus on handling sensitive data at the network edge.}\label{2:classification}
			\vspace{-5mm}
		\end{center}
	\end{figure*}
	
	\indent The realization of trustworthy EI needs to satisfy the four key characteristics of security, reliability, transparency, and sustainability. In order to ensure these trustworthy characterizations in complex edge environments, it is necessary to address the issues described in subsection \uppercase\expandafter{\romannumeral2-C}. Thus, this section aims to provide an in-depth discussion of existing approaches.

	
	\subsection{Solutions for Balanced Security and Privacy Protection}
	

	\indent In this subsection, we focus on discussing solutions for the security issues mentioned in subsection III-C-1 to establish endogenous security capabilities for trustworthy EI systems, as show in Fig. \ref{2:classification}. First, we will investigate security solutions for wireless communications to ensure secure and reliable services in dynamic environments. Second, we will delve into security solutions for learning models to guarantee the integrity and availability of model training and inference in distributed environments. Finally, we will explore privacy-preserving solutions for sensitive data. These solutions aim to strike a balance between security and user experience, avoiding wasted resources and reduced system flexibility. 
	

	\subsubsection{Security Solutions for Wireless Communications}
	
	\indent Researchers usually adopt strategies such as enhanced encryption protocols and secure communication channels to ensure the confidentiality and integrity of data transmission. We summarize the related studies in Tab. \ref{tab2:summary}.
	
	\indent \textbf{Lightweight encryption:} Traditional encryption strategies, such as HE, may not be applicable to resource-constrained edge devices due to their computational requirements~\cite{10044183}. Authors in~\cite{8964439} enhance the security of EI by developing a lightweight and leak-resistant certified key exchange protocol. Meanwhile, the proposed protocol is designed for mainstreamed communication standards. Authors in~\cite{10313321} propose an Attribute-Based Multi-Authority Signed Confidentiality (ABSC) scheme for IoT data sharing. This scheme offloads most of the computation to edge servers, reduces communication and storage costs through short and constant-size ciphertexts, and employs a hierarchical multi-privilege architecture. Unlike the above studies, authors in~\cite{10207065} utilize Quantum Key Distribution (QKD) to protect confidential data for Semantic Information Communication (SIC). In order to reduce the cost of edge users, they use a stochastic planning model to optimize resource allocation and use Shapley values to distribute the cost among QKD service providers.
	

	\indent \textbf{Secure communication channels:} PLS is a classic method, unlike encryption, it does not face the challenges of complex key management and high-overhead key distribution. However, considering constraints such as latency, power consumption, and QoS, security policies based on PLS require trade-offs between system performance and security capabilities. 
	
	\indent Authors in~\cite{wang2023latency} utilize devices that do not participate in Federated Learning (FL), such as Sensor Nodes (SNs), to send jamming signals to defend against eavesdropping attacks. They optimize local training time, model upload time, and transmission power to obtain the best pairing of training nodes and SNs. Considering that friendly jammers have limited power, eavesdroppers are not able to eavesdrop on all channels. Therefore, authors in~\cite{9980450} propose a cooperative jamming power allocation strategy based on game theory that considers eavesdroppers as strategic players.

	
	\indent Unlike PLS, Intelligent Reflective Surface (IRS) technology can support high-dimensional data transmission while enhancing the security of wireless communication networks by adjusting signal amplitudes and phases. Authors in~\cite{9206080} propose a Deep Reinforcement Learning (DRL)-based approach to jointly optimize the beam-forming matrix at the base station and the reflected beam-forming matrix at the IRS. It aims to maximize secure communication to legitimate users in a dynamic environment while meeting QoS requirements. With the advancement of technology, some intelligent attackers are able to selectively employ eavesdropping or active jamming attacks~\cite{li2022reinforcement}. Thus, authors in~\cite{li2022reinforcement} consider a non-cooperative game between a base station and an intelligent attacker. They jointly optimize power allocation and beamforming to improve the secrecy rate. Meanwhile, RL is used to predict the attack behavior and selectively send artificial noise signals.
	
	\begin{table*}[tbp]
		\caption{Security solutions for wireless communications.}
		\label{routing}
		\centering
		\begin{tabular}{|p{20mm}|c|p{85mm}|p{30mm}|} 
			\hline

			\multirow{2}*{\parbox[c]{20mm}{\centering\textbf{Solutions}}} &
			\multirow{2}*{\parbox[c]{5mm}{\centering\textbf{Ref.}}} &
			\multirow{2}*{\parbox[c]{85mm}{\centering \textbf{Description}}} &
			\multirow{2}*{\parbox[c]{30mm}{\centering \textbf{Optimization Metrics}}}
			
			\\

			\hline
			
			\multirow{8}*{\parbox[c]{20mm}{\centering Lightweight encryption}}&

			{\centering ~\cite{8964439}} &

			{\parbox[c][10mm]{85mm}{ A lightweight authentication key agreement to enhance EI security.}}&
			
			
			\parbox[c][10mm]{30mm} {\centering Security and computation costs} 
			\\

			\cline{2-4}
			&{\centering ~\cite{10313321}} &
			
			{\parbox[c][10mm]{85mm}{A lightweight ABSC scheme to ensure secure data sharing in IoT.}}&
			\parbox[c][10mm]{30mm} {\centering Security and computation costs} 
			\\
			
			\cline{2-4}
			&{\centering ~\cite{10207065}} &
			
			{\parbox[c][10mm]{85mm}{A two-stage stochastic optimization model for low-cost QKD-SIC.}}&
			
			\parbox[c][10mm]{30mm}{\centering Security, computation and commutation costs} 
			\\
			
			\hline
			
			\multirow{14}*{\parbox[c]{20mm}{\centering Secure communication channels}}&

			{\centering ~\cite{wang2023latency}} &

			{\parbox[c][10mm]{85mm}{A channel sharing scheme to improve secure FL by incentivizing idle devices to send jamming signals.}}&
			
			{\parbox[c][10mm]{30mm}{\centering Security and FL latency}}  
			\\
			
			\cline{2-4}
			&{\centering ~\cite{9980450}} &
			
			{\parbox[c][10mm]{85mm}{An iterative power allocation strategy to defend against eavesdroppers.}}&
			\parbox[c][10mm]{30mm} {\centering Security} 
			\\
			
			\cline{2-4}
			&{\centering ~\cite{9206080}} &
			
			{\parbox[c][10mm]{85mm}{A DRL-based beamforming method for IRS-assisted secure communication.}}&
			\parbox[c][10mm]{30mm} {\centering Security and QoS} 
			\\
			
			\cline{2-4}
			&{\centering ~\cite{li2022reinforcement}}&
			
			{\parbox[c][10mm]{85mm}{A DQN-based approach that combines IRS and RL to counter intelligent attackers.}}&
			\parbox[c][10mm]{30mm} {\centering Security} 
			\\
			
			\cline{2-4}
			&{\centering~\cite{9674849}}&
			
			{\parbox[c][10mm]{85mm}{An active defense method that uploads confidential information while sending "fake" but meaningful information to confuse the eavesdropper.}}&
			\parbox[c][10mm]{30mm} {\centering Security} 
			\\
			\hline

			\multirow{12}*{\parbox[c]{20mm}{\centering Identity authentication}}&

			{\centering~\cite{10214084}} &

			{\parbox[c][10mm]{85mm}{A collaborative FL-based authentication method for endogenous system security.}}&
			
			{\parbox[c][10mm]{30mm}{\centering Security, privacy, and commutation costs}}  
			\\
			
			\cline{2-4}
			&{\centering~\cite{10089845}} &
			
			{\parbox[c][10mm]{85mm}{A new cooperative authentication scheme helps service providers to authenticate subscribers in a distributed manner.}}&
			\parbox[c][10mm]{30mm} {\centering Security and latency} 
			\\
			
			\cline{2-4}
			&{\centering ~\cite{9963679}} &
			
			{\parbox[c][10mm]{85mm}{An efficient blockchain-based anonymous cross-domain authentication scheme for reliable communications among cross-domain IoT devices.}}&
			\parbox[c][10mm]{30mm} {\centering Security and commutation costs} 
			\\
			
			\cline{2-4}
			&{\centering ~\cite{9919165}}&
			
			{\parbox[c][10mm]{85mm}{A blockchain-based secure and privacy-preserving authentication scheme.}}&
			\parbox[c][10mm]{30mm} {\centering Security, privacy, and commutation costs} 
			\\
			
			\cline{2-4}
			&{\centering ~\cite{10000958}}&
			
			{\parbox[c][10mm]{85mm}{A zero-trust and EI based continuous authentication method for satellite networks.}}&
			\parbox[c][10mm]{30mm} {\centering Security and accuracy} 
			\\

			\hline
		\end{tabular}
		\label{tab2:summary}%
		\vspace{-5mm}
	\end{table*}

	\indent Differently, authors in~\cite{9674849} utilize cyber deception to send true information to the intended receiver while injecting fake information to confuse eavesdroppers. Traps are strategically deployed to attract eavesdroppers and provide them with increasingly clear fake messages, thus establishing a secure
	communication channel between senders and receivers. This method ensures security of exchanged information, even if eavesdroppers gain access to the secret channel information.

	\indent \textbf{Identity authentication:} Due to the increase in the number of wireless users and the openness of electromagnetic wave propagation, wireless communications are vulnerable to security threats. Therefore, ensuring the legitimacy of identities of communicating parties is essential for preventing various attacks and securing the content of communications~\cite{10089845}.

	\indent To accommodate the high dynamics and resource constraints of edge networks, collaborative approaches can provide accurate and robust authentication by sharing multi-dimensional information among devices. For example, authors in~\cite{10214084} propose a collaborative physical layer authentication method based on FL. The scheme utilizes a set of reputable edge devices to co-construct the authenticator to eliminate the limitations of insufficient resources and high authentication complexity. At the same time, the scheme collects multi-dimensional channel state information from different collaborating devices, making it difficult for external adversaries to simulate, thus improving the endogenous security of the system. Similarly, the work in~\cite{10089845} assists service providers in verifying user identities by collecting and processing user location-related characteristics, such as received signal strength and movement trajectories. This article also proposes a context-aware group updating algorithm for adaptively updating cooperative peers and authentication features in dynamic networks.
	

	\indent Blockchain-based authentication is also a new way of identity authentication~\cite{10217209}. Authors in~\cite{9963679} design an efficient blockchain-based anonymous cross-domain authentication scheme to enable reliable communications among cross-domain IoT devices. By combining blockchain and dynamic accumulator technology, the scheme achieves fast authentication, reducing computational pressure for edge devices while ensuring device anonymity. Similarly, authors in~\cite{9919165} propose a complete cross-domain authentication and privacy protection scheme based on federated blockchain. The process of cross-domain authentication is divided into three phases: authorization, on-demand pseudo-identity generation, and identity authentication. In the intra-domain authentication phase, it performs according to the original scheme adopted by the domain. This not only provides excellent scalability, but also greatly reduces the deployment cost and time overhead of the system.
	
	\indent Unlike the above studies, the zero-trust framework aims to provide continuous authentication. Authors in~\cite{10000958} propose a zero-trust and EI-based scheme. Proactive and continuous authentication is achieved by periodically monitoring and re-evaluating variable attributes throughout the request lifecycle. In addition, the article employs an edge intelligence algorithm based on neural-supported decision trees to improve authentication accuracy.
	\begin{table*}[tbp]
		\caption{Security solutions for learning models.}
		\label{routing}
		\centering
		\begin{tabular}{|p{20mm}|c|p{85mm}|p{30mm}|} 
			\hline

			\multirow{2}*{\parbox[c]{20mm}{\centering\textbf{Solutions}}} &
			\multirow{2}*{\parbox[c]{5mm}{\centering\textbf{Ref.}}} &
			\multirow{2}*{\parbox[c]{85mm}{\centering \textbf{Description}}} &
			\multirow{2}*{\parbox[c]{30mm}{\centering \textbf{Optimization Metrics}}}
			
			\\

			\hline
			
			\multirow{4}*{\parbox[c]{20mm}{\centering Anomaly detection}}&

			{\centering~\cite{shi2021federated}} &

			{\parbox[c][10mm]{85mm}{A federal anomaly analysis framework to defend against local model poisoning attacks
					in distributed ML frameworks.}}&
			
			
			\parbox[c][10mm]{30mm} {\centering Security, accuracy, and computation costs} 
			\\

			\cline{2-4}
			&{\centering ~\cite{zhou2022differentially}} &
			
			{\parbox[c][10mm]{85mm}{A FL model combining anomaly detection and DP to resist poisoning attacks and protect
					privacy.}}&
			\parbox[c][10mm]{30mm} {\centering Security, privacy, and accuracy} 
			\\
			
			\hline
			
			\multirow{18}*{\parbox[c]{20mm}{\centering Robust aggregation}}&

			{\centering ~\cite{zhang2022lsfl}} &

			{\parbox[c][10mm]{85mm}{A novel lightweight privacy-preserving crowdsourced FL scheme to support secure model aggregation.}}&
			
			{\parbox[c][10mm]{30mm}{\centering Security, privacy, and computation costs}}  
			\\
			
			\cline{2-4}
			&{\centering ~\cite{10250861}} &
			
			{\parbox[c][10mm]{85mm}{A layered privacy-preserving defense architecture against poisoning attacks in data heterogeneity scenarios.}}&
			\parbox[c][10mm]{30mm} {\centering Security, privacy, accuracy, and communication costs} 
			\\
				\cline{2-4}
			&{\centering ~\cite{zhao2021sear}} &
			
			{\parbox[c][10mm]{85mm}{An efficient TEE-based aggregation framework for Byzantine robust FL.}}&
			\parbox[c][10mm]{30mm} {\centering Security and accuracy} 
			\\
			
			\cline{2-4}
			&{\centering ~\cite{li2021lomar}} &
			
			{\parbox[c][10mm]{85mm}{A two-stage defense algorithm to analyze local feature patterns of malicious remote updates.}}&
			\parbox[c][10mm]{30mm} {\centering Security and accuracy} 
			\\
			
			\cline{2-4}
			&{\centering ~\cite{elkordy2022basil}}&
			
			{\parbox[c][10mm]{85mm}{A fast and efficient byzantine robustness algorithm for P2P systems.}}&
			\parbox[c][10mm]{30mm} {\centering Security, efficiency, and accuracy} 
			\\
			
			\cline{2-4}
			&{\centering\cite{9916263}}&
			
			{\parbox[c][10mm]{85mm}{A blockchain consensus-based approach to defend against model poisoning attacks.}}&
			\parbox[c][10mm]{30mm} {\centering Security and latency} 
			\\
			
			\cline{2-4}
			&{\centering \cite{9761745}}&
			
			{\parbox[c][10mm]{85mm}{A secure and privacy-preserving decentralized learning system.}}&
			\parbox[c][10mm]{30mm} {\centering Security, privacy, and efficiency} 
			\\
			\hline

			{\parbox[c]{20mm}{\centering Adversarial training}}&

			{\centering ~\cite{qian2022ei}} &

			{\parbox[c][10mm]{85mm}{A dynamic defense mechanism to improve EI classification accuracy in adversarial
					settings.}}&
			
			{\parbox[c][10mm]{30mm}{\centering Security and accuracy}}  
			\\
			\hline

			{\parbox[c][10mm]{20mm}{\centering GAN}}&

			{\centering ~\cite{8892628}} &

			{\parbox[c][10mm]{85mm}{A decentralized fast vigilance framework for identifying adversarial attacks in IAISs.}}&
			
			{\parbox[c][10mm]{30mm}{\centering Security and latency}} \\
			\hline

		\end{tabular}
		\label{tab3:summary}%
		\vspace{-5mm}
	\end{table*}

	\subsubsection{Security Solutions for Learning Models}
	
	\indent In EI, poisoning attacks and adversarial attacks are prevalent methods of attack. In response to these threats, we provide an overview of the primary defense methods against each of these attack types. In Tab. \ref{tab3:summary}, we briefly summarize the solutions to achieve EI model security.

	\indent \textbf{Defense strategies for poisoning attacks:} Poisoning attacks cause damage to the reliability and usability of models by manipulating and injecting poisoned data. In EI, FL is a commonly used approach for model training, where the adversary usually directly poisons local updates without modifying the training data. Unlike centralized learning frameworks, defenses against poisoning attacks in FL need to take into account that local training data is not visible to the public. One defense idea is to detect malicious clients by distinguishing between toxic and benign updates.

	\indent Authors in~\cite{shi2021federated} develop an unsupervised anomaly detection method based on Support Vector Machines (SVM). They introduce a separate validation operation for each potentially malicious local model, to improve anomaly detection accuracy. Differently, authors in~\cite{zhou2022differentially} propose a weight-based detection scheme. It provides edge nodes with small validation datasets to detect and filter anomalous parameters uploaded by malicious end devices. Based on detection results, edge nodes set appropriate parameter weights to eliminate the effect of pseudo-parameters on the model. In addition, they use DP techniques to provide privacy-preserving measures for sensitive data.

	\indent Another approach is to enforce robust aggregation algorithm to counter interference from unknown adversaries. Authors in~\cite{10250861} propose a secure cosine similarity scheme to identify toxic gradients using HE as the underlying technique. The communication overhead is also reduced by replacing the remote communication method with intra-cluster communication. Authors in~\cite{zhang2022lsfl} design a lightweight dual-server secure aggregation protocol that utilizes a third-party server to compute the distance between the model updates from different participants and the average model update. Based on the deviation degrees, the protocol selects the top k nearest participants, and their local updates are considered benign. Authors in~\cite{zhao2021sear} use the Euclidean distance of the model to filter malicious updates and subsequently compute the median of the remaining model coordinates to ensure the accuracy of results. In addition, they provide a secure aggregation environment with the help of TEE (i.e., Intel SGX) for global model security. Differently, authors in~\cite{li2021lomar} propose a scoring model that employs kernel density estimation to evaluate updates from remote clients. They statistically approximate the optimal threshold to distinguish malicious updates from clean ones. 
	
	\indent Instead of relying on a central parameter server, authors in~\cite{elkordy2022basil} propose a fast and computationally efficient byzantine-robust algorithm for fully decentralized training systems. Their algorithm utilizes a new sequential, memory-assisted, and performance-based criterion for training on logical rings while filtering out byzantine users.  Similarly, researchers in~\cite{9916263} and~\cite{9761745} also focus on robust aggregation for fully decentralized training systems. The difference from~\cite{elkordy2022basil} is that they use blockchain to provide a transparent and secure process.
	\begin{table*}[tbp]
		\caption{Privacy-preserving solutions for sensitive data.}
		\label{routing}
		\centering
		\begin{tabular}{|p{20mm}|c|p{85mm}|p{30mm}|} 
			\hline

			\multirow{2}*{\parbox[c]{20mm}{\centering\textbf{Solutions}}} &
			\multirow{2}*{\parbox[c]{5mm}{\centering\textbf{Ref.}}} &
			\multirow{2}*{\parbox[c]{85mm}{\centering \textbf{Description}}} &
			\multirow{2}*{\parbox[c]{30mm}{\centering \textbf{Optimization Metrics}}}
			
			\\

			\hline
			
			\multirow{6}*{\parbox[c]{20mm}{\centering Lightweight encryption}}&

			{\centering~\cite{9663203}} &

			{\parbox[c][10mm]{85mm}{A crypto-neural network inference system at the network edge.}}&
			
			
			\parbox[c][10mm]{30mm} {\centering Privacy and latency} 
			\\
			\cline{2-4}
			
			&{\centering~\cite{8700229}} &

			{\parbox[c][10mm]{85mm}{A novel lightweight framework to address privacy issues for mobile sensing.}}&
			
			
			\parbox[c][10mm]{30mm} {\centering Privacy, latency, and computation costs} 
			\\

			\cline{2-4}
			
			&{\centering~\cite{10271394}} &

			{\parbox[c][10mm]{85mm}{ A lightweight certificate-free multi-user encryption algorithm for mobile devices.}}&
			
			
			\parbox[c][10mm]{30mm} {\centering Privacy, communication and computation costs} 
			\\
			\hline
			
			\multirow{6}*{\parbox[c]{20mm}{\centering Data desensitization}}&

			{\centering ~\cite{10098897}} &

			{\parbox[c][10mm]{85mm}{A novel hybrid learning approach to enhance privacy protection by extracting special features from sensitive data.}}&
			
			{\parbox[c][10mm]{30mm}{\centering Privacy and latency}}  
			\\
			
			\cline{2-4}
			&{\centering \cite{10043027}} &
			
			{\parbox[c][10mm]{85mm}{A two-level DP mechanism to enhance privacy protection in cloud-edge-end layered FL frameworks.}}&
			\parbox[c][10mm]{30mm} {\centering Privacy, accuracy, and latency} 
			\\

			\cline{2-4}
			&{\centering ~\cite{9750858}} &
			
			{\parbox[c][10mm]{85mm}{A collaborative approach to enhance privacy preservation of inference processes.}}&
			\parbox[c][10mm]{30mm} {\centering Privacy and latency} 
			\\
			
			\hline

			\multirow{8}*{\parbox[c]{20mm}{\centering Surrogate data}}&

			{\centering \cite{shejwalkar2021membership}} &

			{\parbox[c][10mm]{85mm}{A knowledge transfer-based method to trade off data privacy and classification accuracy.}}&
			
			{\parbox[c][10mm]{30mm}{\centering Privacy and accuracy}}  
			\\
			
			\cline{2-4}
			&{\centering ~\cite{hu2022defending}} &
			
			{\parbox[c][10mm]{85mm}{A GAN method to generate new data, thereby preserving the privacy of sensitive training
					data.}}&
			\parbox[c][10mm]{30mm} {\centering Privacy and accuracy} 
			\\
			
			\cline{2-4}
			&{\centering ~\cite{9479764}} &
			
			{\parbox[c][10mm]{85mm}{A novel adversarial sample generation method based on the Firefly algorithm for data protection.}}&
			\parbox[c][10mm]{30mm} {\centering Privacy and latency} 
			\\
			\hline
			
		\end{tabular}
		\label{tab4:summary}%
		\vspace{-5mm}
	\end{table*}

	\indent \textbf{Defense strategies for adversarial attacks:} Adversary attacks aim to deceive and mislead the output of a model by making a minor but intentionally designed modification to the input data. Compared to large-scale cloud-based models, compression models running on edge devices typically have fewer parameters and computational resources, and adversarial examples can easily mislead them~\cite{qian2022ei}. Adversarial training enhances the ability to handle attacks by incorporating adversarial examples into the training data. However, these defense measures rely on computationally expensive solutions to generate effective adversarial examples, limiting their applicability in EI. 
	
	\indent  To achieve lightweight defense, authors in~\cite{qian2022ei} propose a dynamic defense mechanism that combines techniques such as KD, MTD, and Bayesian Stackelberg games to improve model robustness, especially the classification accuracy of EI in an adversarial setting. Authors in~\cite{8892628} focus on adversarial attacks in Industrial AI Systems (IAISs). The proposed method combines multiple DL models and multiple Conditional Generative Adversarial Networks (CGAN). The CGAN control plane and the DL model data plane operate without the need for complex robustness reinforcement of the original DL model. Author in~\cite{xu2022lancex} introduce LanCeX, which is designed to counter compression adversarial attacks in embedded recognition scenarios. LanCeX comprises a detection phase to identify potential adversarial patterns and a data recovery method to mitigate adversarial perturbations in the input data. The proposed defense approach is both universal and lightweight, making it suitable for resource-constrained embedded systems.

	\subsubsection{Privacy-preserving Solutions for Sensitive Data}
	
	\indent Researchers develop various privacy-preserving schemes aimed at balancing system performance and user privacy protection, providing a solid foundation for reliability and user experience in EI. As shown in Tab. \ref{tab4:summary}, we briefly summarize some solutions.
	
	\indent Authors in~\cite{9663203} propose a lightweight encrypted edge inference systems. They use binary neural networks to reduce resource requirements for deploying models on edge devices, while using secret sharing techniques to provide privacy guarantees. Similarly, authors in~\cite{8700229} use secret sharing based encryption to address privacy-preserving CNN feature extraction for mobile sensing, while significantly reducing the latency and overhead of the end device. Differently, authors in~\cite{10271394} propose a new matching encryption method that provides bilateral access control between the sender and the receiver. The proposed scheme employs a low-power pair-free operation and supports a one-to-many setup to avoid encrypting each receiver's message individually. In addition, they introduce a certificate-less encryption method to address the key escrow problem.
	
	\indent In addition, the risk of privacy leakage is reduced by employing desensitization techniques to process sensitive information, which can be achieved by DP. Authors in~\cite{10098897} propose a privacy-preserving and latency-aware DL framework that addresses privacy and latency issues in EI systems. It introduces inductive learning and a new local DP algorithm that allows edge devices to apply random noise to features extracted from sensitive data before transmitting them to a central server. Authors in \cite{10043027} utilize a cloud-edge-end hierarchical FL framework to offload the training burden of the device to the nearest edge to improve efficiency, while protecting privacy with a two-level DP mechanism. Differently, authors in~\cite{9750858} establish a collaborative strategy  to hide the attributes of the original inputs by distributing CNN feature mappings across multiple heterogeneous IoT devices, thus making it impossible for malicious devices to recover the original data. At the same time, a trade-off between latency and privacy is established.

	\begin{figure*}[htbp]
		\begin{center}
			\includegraphics[scale = 0.5]{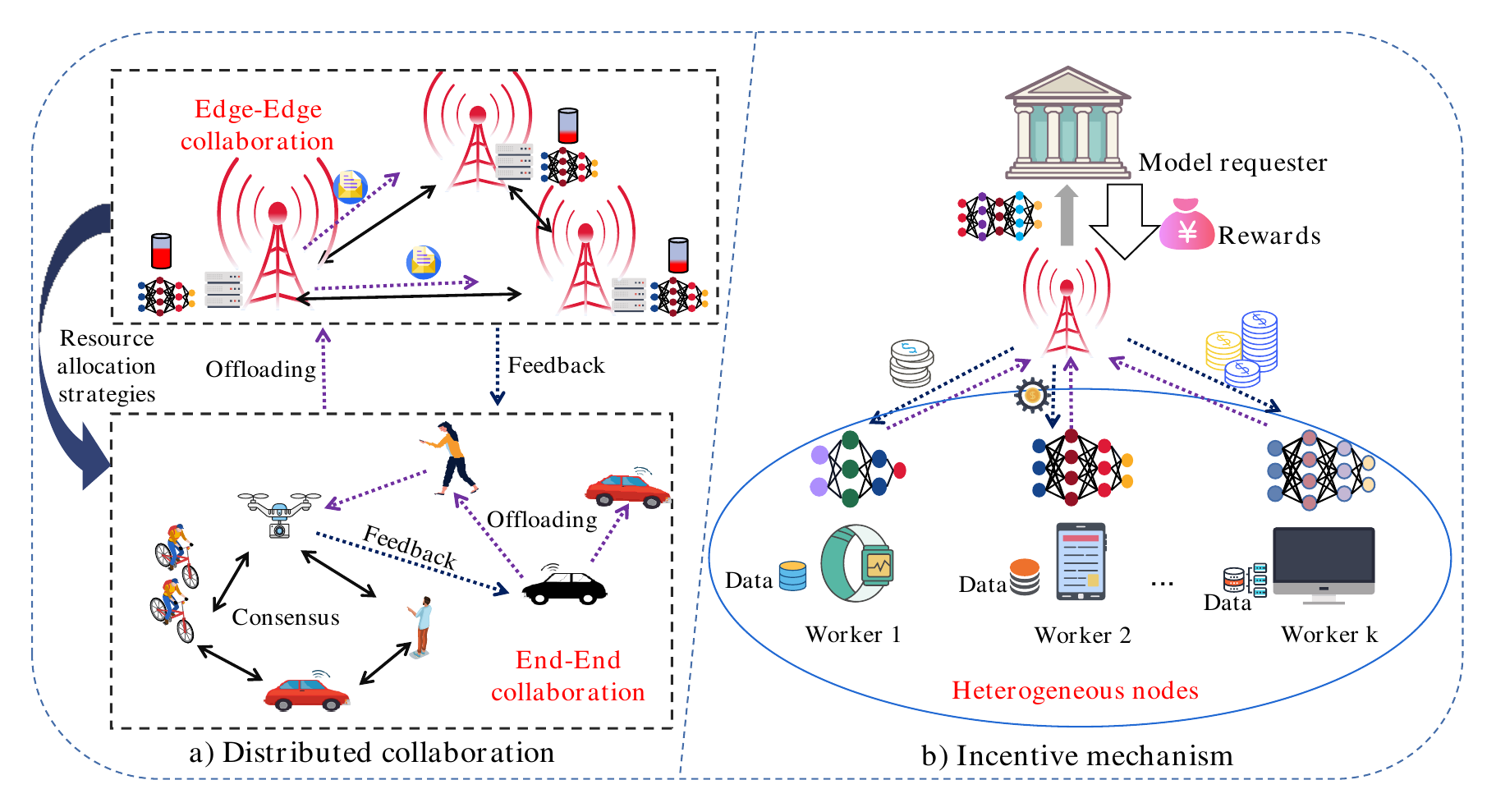}\\
			\caption{Illustrative solutions of reliability: a) Distributed collaboration: It allows heterogeneous devices to work together and share knowledge, reducing the risk of single points of failure while increasing resource utilization; and b) Incentive mechanisms: Rewards are given for honest and high-quality resource contributions. }\label{5:classification}
			\vspace{-5mm}
		\end{center}
	\end{figure*}

	\indent Privacy protection also can be achieved by replacing sensitive data with similarly distributed proxy data. Authors in~\cite{huang2021damia} propose a defense method, where the model is trained on a different but related dataset, avoiding direct access to original sensitive dataset. Authors in \cite{shejwalkar2021membership} utilize an unprotected model trained on private data and transfer its knowledge to a student model trained on labeled reference data. Authors in~\cite{hu2022defending} employ GANs to generate new data that required for training. To enhance quality of the generated data by GANs, authors utilize truncation techniques and clustering algorithms during the generation process for different types of data. Similarly, authors in~\cite{9479764} propose a data protection method based on data disturbance and adversarial training. They also introduce a new adversarial sample generation method based on firefly algorithm.
	
	\indent \textbf{ Lesson 1:} In this subsection, we delve into three aspects of security for edge communication, security for learning models, and privacy protection, providing flexible and balanced security strategies for resource-constrained edge networks. A comprehensive analysis of these solutions reveals several valuable lessons: \textit{\romannumeral1}) In terms of security for edge communications, the use of lightweight encryption and secure communication channels is the key to improve data confidentiality and integrity. Novel authentication methods, such as collaborative authentication, excel in adapting to highly dynamic edge networks and resource limitations; \textit{\romannumeral2}) Methods such as unsupervised anomaly detection and robust aggregation algorithms are effective in detecting and filtering malicious updates. The use of dynamic defense mechanisms, knowledge distillation, and multi-task learning helps to improve model robustness; \textit{\romannumeral3}) Methods such as DP, feature extraction and GAN are effective in protecting data privacy while reducing the required computation and communication resources.
	
	\indent However, the trade-off between security and resource efficiency needs to be further optimized, especially given the limitations of edge devices. Additionally, the evolving landscape of cyber threats necessitates adaptive and proactive security strategies. Striking the right balance between security measures and usability remains a persistent challenge in the dynamic context of EI. 
	

	\subsection{Solutions for Reliability}
	
	\indent As discussed in subsection II-C-2, highly dynamic edge environments where the addition of new nodes may change the network connectivity and the departure of nodes may lead to insufficient resources in certain areas of the network. Such topological changes pose challenges to the network stability and reliability. Therefore, the system requires effective measures to adapt to the rapidly changing network environment to ensure user experience, task execution, and system functionality. In this subsection, we detail solutions, including distributed collaboration and incentive mechanisms, to address the issues and thus establish reliability of trustworthy EI, which is shown in Fig. \ref{5:classification}.

	
	\subsubsection{Distributed Collaboration}
	
	\begin{table*}[tbp]
		\caption{Distributed collaboration solutions for reliability.}
		\label{routing}
		\centering
		\begin{tabular}{|p{20mm}|c|p{75mm}|p{30mm}|p{30mm}|} 
			\hline

			\multirow{2}*{\parbox[c]{20mm}{\centering\textbf{Solutions}}} &
			\multirow{2}*{\parbox[c]{5mm}{\centering\textbf{Ref.}}} &
			\multirow{2}*{\parbox[c]{75mm}{\centering \textbf{Description}}} &
			\multirow{2}*{\parbox[c]{30mm}{\centering \textbf{Network Scenarios}}} &
			\multirow{2}*{\parbox[c]{30mm}{\centering \textbf{Optimization Metrics}}}
			
			\\

			\hline
			
			\multirow{12}*{\parbox[c]{20mm}{\centering Autonomous decision-making}}&

			{\centering ~\cite{9994770}} &

			{\parbox[c][10mm]{75mm}{A multi-UAV collaborative system combining multi-agent RL and attention mechanisms for efficient resource allocation.}}&
			
			
			\parbox[c][10mm]{30mm} {\centering  Multi-UAV assisted vehicular networks} &
			\parbox[c][10mm]{30mm} {\centering Latency and communication costs}
			\\
			
			\cline{2-5}
			&{\centering \cite{10026241}} &
			
			{\parbox[c][10mm]{75mm}{A lightweight imitation learning-based distributed agent strategy for autonomous cooperation.}}&
			\parbox[c][10mm]{30mm} {\centering Edge networks} &
			\parbox[c][10mm]{30mm} {\centering Latency} 
			\\
			
			\cline{2-5}
			&{\centering \cite{xiao2022perception}} &
			
			{\parbox[c][10mm]{75mm}{A perceptual task offloading framework to improve the reliability of autonomous driving.}}&
			\parbox[c][10mm]{30mm} {\centering IoV} &
			\parbox[c][10mm]{30mm} {\centering Latency} 
			\\
			\cline{2-5}
			&{\centering~\cite{8698845}} &
			
			{\parbox[c][10mm]{75mm}{An intelligent resource allocation framework based on multi-task deep RL algorithm.}}&
			\parbox[c][10mm]{30mm} {\centering  IoT } &
			\parbox[c][10mm]{30mm} {\centering Latency and power consumption} 
			\\
			\cline{2-5}
			&{\centering~\cite{9769868}} &
			
			{\parbox[c][10mm]{75mm}{A novel multi-exit DNN inference acceleration framework.}}&
			\parbox[c][10mm]{30mm} {\centering Edge inference networks} &
			\parbox[c][10mm]{30mm} {\centering Latency and computation costs} 
			\\
			\hline
			
			\multirow{6}*{\parbox[c]{20mm}{\centering Knowledge and model sharing}}&

			{\centering ~\cite{9562522}} &

			{\parbox[c][10mm]{75mm}{A semi-decentralized learning architecture combining device-to-server and D2D communication paradigms.}}&
			
			{\parbox[c][10mm]{30mm}{\centering Edge training networks}}  &
			{\parbox[c][10mm]{30mm}{\centering Latency, accuracy and energy consumption}} 
			\\
			
			\cline{2-5}
			&{\centering ~\cite{9928395}} &
			
			{\parbox[c][10mm]{75mm}{A FL framework for edge server collaboration based on decentralized consensus.}}&
			\parbox[c][10mm]{30mm} {\centering Edge training network}&
			\parbox[c][10mm]{30mm} {\centering Latency and accuracy}
			\\
			\cline{2-5}
			&{\centering ~\cite{deb2022loop}} &
			
			{\parbox[c][10mm]{75mm}{A fragmented distributed learning approach for IoT devices.}}&
			\parbox[c][10mm]{30mm} {\centering  IoT }&
			\parbox[c][10mm]{30mm} {\centering Latency, computation costs, and energy consumption}
			\\
			
			\hline

			\multirow{8}*{\parbox[c]{20mm}{\centering Decentralized collaboration}}&

			{\centering ~\cite{hwang2022decentralized}} &

			{\parbox[c][10mm]{75mm}{A decentralized DRL-based resource allocation and task scheduling approach.}}&
			
			{\parbox[c][10mm]{30mm}{\centering UAV-MEC networks}}&  
			{\parbox[c][10mm]{30mm}{\centering Latency and energy consumption}}
			\\
			
			\cline{2-5}
			&{\centering ~\cite{10339164}} &
			
			{\parbox[c][10mm]{75mm}{A collaborative DRL approach for resource allocation that ensures long-term energy savings.}}&
			\parbox[c][10mm]{30mm} {\centering MEC networks} &
			\parbox[c][10mm]{30mm} {\centering Energy efficiency}
			\\
			
			\cline{2-5}
			&{\centering ~\cite{10141684}} &
			
			{\parbox[c][10mm]{75mm}{An efficient decentralized GNN-based training algorithm.}}&
			\parbox[c][10mm]{30mm} {\centering Decentralized communication networks} &
			\parbox[c][10mm]{30mm} {\centering Communication costs} 
			\\
			\hline
			
		\end{tabular}
		\label{tab5:summary}%
		\vspace{-5mm}
	\end{table*}

	\indent Through distributed collaboration, the EI system is able to build resilient and adaptable characteristics. As shown in Tab.\ref{tab5:summary}, we summarize the relevant literature on distributed collaborative EI solutions to achieve reliability, which include: autonomous decision-making, knowledge and model sharing, and decentralized mechanisms.
	
	\indent First, autonomous decision-making assists edge devices in adjusting their behavior to adapt to dynamic environments and task requirements. For instance, authors in~\cite{9994770} focus on resource allocation of multiple UAVs in a ground vehicle network. By using multi-agent RL algorithm, each UAV acts as an intelligent agent with centralized training and distributed execution to collaborate on resource. In addition, an attention mechanism is introduced where agents can further optimize their local models based on information from other agents. Authors in~\cite{10026241} propose an imitation learning based strategy to train distributed agents. Unlike RL, which requires a specific reward function, the imitation learning supports to teach complex tasks to agent with few information, thus turning the training process into fitting an expert demonstration distribution. Authors in~\cite{xiao2022perception} propose a multi-level perceptual task offloading framework, in which a vehicle is able to utilize other nearby AVs and Roadside Units (RSUs) to perform collaborative computation for integrated perception of the region of interest. In the proposed approach, the AVs are able to dynamically assign, offload, and execute tasks based on the characteristics and latency requirements of sensing tasks. 
	
	\indent In addition, collaborative resource allocation among nodes is a key solution to achieve network reliability and improve QoS. Authors in~\cite{8698845} propose an intelligent resource allocation framework, which trains a DNN to predict resource allocation behavior through self-supervised learning. Subsequently, action prediction is combined with multi-task learning to enhance the performance of DNN. Differently, authors in~\cite{9769868} consider the synergy of computational resources from mobile devices and edge servers to accelerate multi-outlet DNN inference. In the proposed model, bi-directional dynamic programming is used for exit point selection, and then DRL is used to model partitioning and resource allocation strategies.
	
	
	\indent Second, the overall performance is improved by sharing knowledge, models and experience among collaborative nodes. Authors in~\cite{10043782} present a theoretical framework for real-time evolutionary learning of states in decentralized edge inference network. Specifically, the goal of each agent is to infer a time-varying state in a decentralized manner by using its local observations and messages from other nodes within the communication range. The article tries to provide a communication-efficient coding strategy for generating transmission messages and a sufficient condition for the boundedness of the distributed inference error over time for all agents. In~\cite{9562522}, authors introduce a consensus mechanism that utilizes Device-to-Device (D2D) communication to mitigate model disagreement and improve resource efficiency. This work can be considered as a learning approach in the middle star topology between traditional FL and fully decentralized architectures, resulting in a new semi-decentralized learning architecture. Similarly, authors in~\cite{9928395} devise a joint optimization method to facilitate collaborative learning among a large number of edge devices in a wide region. By utilizing multiple aggregators to mitigate concerns about single point of failure, the approach is more scalable than the traditional FL framework.
	
		\begin{table*}[tbp]
		\caption{Incentivization solutions for reliability.}
		\label{routing}
		\centering
		\begin{tabular}{|p{25mm}|c|c|c|c|c|c|p{100mm}|} 
			\hline

			\multirow{8}*{\parbox[c]{25mm}{\centering\textbf{Categories}}} &	
			\multirow{8}*{\parbox[c]{5mm}{\centering\textbf{Ref.}}} &
			\multicolumn{5}{c|}{\parbox[c]{20mm}{\centering \textbf{Solutions}}} &
			\multirow{8}*{\parbox[c]{100mm}{\centering\textbf{Description}}}\\
			
			\cline{3-7}
			&&
			
			\parbox[t]{1mm}{\multirow{7}*{\rotatebox[origin=c]{90}{\centering \textbf{Auction Theory}}}}	&
			\parbox[t]{1mm}{\multirow{7}*{\rotatebox[origin=c]{90}{\centering \textbf{Game Theory}}}} &		
			\parbox[t]{1mm}{\multirow{7}*{\rotatebox[origin=c]{90}{\centering \textbf{Contract Theory}}}} & \parbox[t]{1mm}{\multirow{7}*{\rotatebox[origin=c]{90}{\centering \textbf{Smart Contract}}}} & 
			\parbox[t]{1mm}{\multirow{7}*{\rotatebox[origin=c]{90}{\centering \textbf{DRL}}}}&	\\

			&&&&&&& \\
			&&&&&&& \\
			&&&&&&& \\
			&&&&&&& \\
			&&&&&&& \\	
			&&&&&&& \\

			\hline
			
			
			\multirow{14}*{\parbox[t]{25mm}{\centering Incentivization via node contributions}}&	
			
			{~\cite{deng2022improving}}&	{$\surd$}	& {$\times$}& {$\times$}	& {$\times$}  & {$\times$}  &{\parbox[c][10mm]{100mm}{A framework for integrating quality estimation, reverse auction incentives, and automatic weighted aggregation for high-quality model training.}} \\
			
			\cline{2-8}
			&{\centering ~\cite{ng2021hierarchical}} & {$\surd$} & {$\surd$}& {$\times$}	& {$\times$} & {$\times$}  & {\parbox[c][10mm]{100mm}{A two-tier incentive mechanism that considers local data contribution and energy consumption of training nodes.}} \\
			
			\cline{2-8}
			&{\centering ~\cite{9479786}} & {\centering $\surd$}& {\centering $\surd$}& {\centering $\times$} & {\centering $\times$} &{$\times$}  & {\parbox[c][10mm]{100mm}{A two-tier resource allocation and incentive mechanism for decentralized learning-based systems.}}\\
			
			\cline{2-8}
			&{\centering ~\cite{9155268}} &{\centering $\times$}	& {\centering $\times$}& {\centering $\times$}	& {\centering $\times$}  &{$\surd$}  &{\parbox[c][10mm]{100mm}{A DRL-based automatic pricing strategy to incentivize nodes to contribute computational resources.}} \\
			
			\cline{2-8}
			&{\centering ~\cite{hu2022autofl} } &
			{\centering $\times$}	& 
			{\centering $\surd$}& 
			{\centering $\times$}	& 
			{\centering $\times$}  &
			{$\times$}  &
			{\parbox[c][10mm]{100mm}{A Bayesian game-based incentive mechanism to encourage nodes to contribute their
					data and computational resources in FL.}} \\
			
			\hline
			
			\multirow{8}*{\parbox[c]{25mm}{\centering Incentivization via blockchain}}&	
			
			{\centering ~\cite{chen2022dim}}&	
			{\centering $\times$}	& 
			{\centering $\surd$}& 
			{\centering $\times$}	&
			{\centering $\surd$} &   
			{$\times$}  &
			{\parbox[c][10mm]{100mm}{ A dynamic incentive model combining evolutionary game theory and smart contracts to encourage users to participate in data sharing.}} \\
			
			\cline{2-8}
			&{\centering ~\cite{kang2019incentive}}&	
			{\centering $\times$}	& 
			{\centering $\times$} & 
			{\centering $\surd$} &
			{\centering $\times$} &
			
			{\centering $\times$} &
			{\parbox[c][10mm]{100mm}{ A contract-based incentive mechanism to encourage nodes to participate in training and share resources, and store the reputation of nodes via blockchain.}} \\
			
			\cline{2-8}&
			{\centering ~\cite{wang2022infedge}} &	
			{\centering $\times$}	 & 
			{\centering $\surd$}   & 
			{\centering $\times$}	 & 
			{\centering $\surd$}   &  
			{\centering $\times$} &
			{\parbox[c][10mm]{100mm}{A blockchain-based FL incentive mechanism to balance system overhead and model performance.}} \\
			
			\cline{2-8}&
			{\centering~\cite{wang2023incentive}} &	
			{\centering $\times$}	 & 
			{\centering $\surd$}   & 
			{\centering $\times$}	 & 
			{\centering $\times$}   &  
			{\centering $\times$} &
			{\parbox[c][10mm]{100mm}{A two-stage Stackelberg game approach to optimize resource allocation in blockchain-based FL systems.}} \\
			
			\hline
			
			
			
			

			
			\multicolumn{8}{c}{(``$\surd$'' if the research satisfies the solution, ``$\times$'' if not)}

		\end{tabular}
		\label{tab6:summary}%
		\vspace{-5mm}
		
	\end{table*}

	\indent Different from~\cite{9562522} and~\cite{9928395}, authors in~\cite{deb2022loop} propose fragmented learning in resource-constrained IoT edge devices. The method splits the training operation for each data point into atomic operations and executes them on Fog Nodes (FNs). An iterative implementation of fragmentation learning enables FNs to transfer partially learned weights to the next appropriate FN for further training, and repeats the process until the training is complete. By analyzing main parameters and using a greedy heuristic algorithm, the algorithm selects optimal FNs for the training operation, which tries to reduce the probability of interruptions caused by device failures.
	
	\indent Finally, decentralized mechanisms for decision making and task assignment ensure that nodes can work together without centralized control. Authors in~\cite{hwang2022decentralized} propose a multi-agent DRL approach for decentralized implementation, where multiple UAVs collaborate to determine their computational and communication strategies. Similarly, authors in~\cite{10339164} present a fully decentralized multi-agent DRL algorithm for autonomous resource allocation in heterogeneous mobile edge networks. The algorithm has a multi-actor shared criticism architecture and a regional training distributed execution framework, which aims to stabilize model training and reduce information exchange.
	
	\indent Graph Neural Networks (GNNs) have attracted much academic attention for its ability to efficiently process graph data, adapt to the dynamic nature of wireless networks, and enable decentralized management and control. Authors in~\cite{10141684} propose a personalized training algorithm based on graph attention to address the problems caused by non-independently identically and distributed data in traditional distributed learning. The algorithm allows each agent to train a local model that personalizes data by learning the specific weights of different neighboring nodes without prior knowledge of the graph structure or the data distribution of neighboring nodes. However, the fading of wireless channels makes GNN affected by the information exchange among neighbors~\cite{9606569}. Therefore, authors in~\cite{9606569} try to enhance the robustness of decentralized GNNs in the inference phase through two new re-transmission mechanisms. 
	
	

	\subsubsection{Incentive Mechanisms}
	
	\indent Appropriate incentive mechanisms can inspire edge nodes to actively participate in collaboration and contribute their own resources, which helps to mitigate the impact of node withdrawal, thus improving the overall reliability and adaptability of the system. In the following, we introduce incentive mechanisms based on node contributions and blockchain, which are summarized in Tab. \ref{tab6:summary}.
	
	\indent \textbf{Incentivization via node contributions:} The quality of model updates can vary greatly due to a variety of factors such as training data volume, data quality, and data distribution. As a result, some studies utilize them to measure customer contributions. Authors in~\cite{deng2022improving} employ a reverse auction model to incentivize high-quality and low-cost computing nodes to participate in training process.
	
	\indent A single-layer incentive mechanism may be not efficient in motivating all parties. To address this challenge, researchers in~\cite{ng2021hierarchical} and~\cite{9479786} propose a two-layer incentive mechanism. In the lower layer, the size of reward pool and data allocation of training nodes are determined based on data volume, data quality, and privacy budget provided by each data owner. In the upper layer, profits are allocated to training nodes based on their marginal contributions to the model publisher, i.e., the impact on the global model performance.
	
	\indent Instead of focusing solely on a single dimension, such as node data quality and quantity, which may not guarantee long term system efficiency and stability, some studies address situations where nodes possess multi-dimensional attributes, such as heterogeneity communication resources and computational capabilities~\cite{wang2022infedge}. Authors in~\cite{9155268} discuss incentives for using DRL to balance the trade-off between latency and payment in edge learning. At the beginning of each training round, the aggregator publishes the price of each Edge Node (EN) and the EN determine the computational resources they provide for model training. The incentives are presented in the form of a hierarchical game, using an actor-critic network model to overcome the challenges of the dynamic environment and the access to parameter information of ENs. Similarly, authors in~\cite{hu2022autofl} consider the balance among rewards, costs, edge communication and computing capabilities, allowing nodes to determine their contributions of local data and computational resources in each training round. 
	
	\indent \textbf{ Incentivization via blockchain:} Blockchain-based incentives create a trusted and transparent environment for collaborative EI. It is able to record and preserve the historical reputation of nodes, preventing interference from unreliable participants~\cite{ning2021intelligent}. Authors in~\cite{chen2022dim} combine reputation-based and payment-based incentives, using "credibility coins" as encrypted cryptocurrency for data transactions. At the same time, they introduce a dynamic incentive model based on evolutionary game theory to analyze user interactions and the stability of strategies. Authors in~\cite{kang2019incentive} integrate reputation and contract theory to ensure fair rewards. They utilize blockchain for secure reputation management of training nodes, providing non-repudiation and anti-tampering properties in a decentralized manner. These reputation-based approaches assess the trustworthiness and contributions of participating nodes, either based on their impacts on model performance or through social interactions, to ensure integrity and reliability of learning.
	
	\indent Moreover, authors in~\cite{wang2022infedge} leverage blockchain-based incentive mechanisms to optimize the balance between system overhead and model performance. They achieve this by compensating relevant nodes with different resources, while also enhancing data privacy through mechanisms embedded in smart contracts. This approach establishes a credible, faster, and transparent resource trading system. Similarly, in~\cite{wang2023incentive}, the authors formulate the resource allocation problem within blockchain-based learning as a two-stage Stackelberg game. They assist model owners in reward allocation and clients in determining their computational resources for both model training and task mining.
	
	\indent \textbf{Lesson 2:} With solutions such as distributed collaboration and incentives, it is possible to build an intelligent system at the network edge with a high degree of flexibility and reliability to adapt to dynamic communication environments and collaboration among heterogeneous nodes. This lays the foundation for achieving sustainable QoS. Distributed collaboration enables that nodes to communicate with each other, share knowledge, and tolerate single point of failure. In particular, the combination of blockchain and smart contracts helps to ensure the credibility and transparency of the incentives.
	
	\indent However, when implementing and optimizing distributed collaboration and incentives, it is necessary to ensure fairness among nodes and guard against free-riding behavior. This includes avoiding over-centralization of rewards while ensuring that nodes share resources honestly. In addition, the manageability of distributed and decentralized networks is challenging compared to centralized architectures and requires a combination of technical, institutional and social considerations.

	\subsection{Solutions for Transparency}\label{Solutions for Interpretability}
	
	
	\begin{table*}[tbp]
		\centering
		\caption{\centering Transparency solutions based on decision interpretability.}
		\begin{tabular}{|p{20mm}|c|p{100mm}|p{20mm}|}
			\hline

			\multirow{2}*{\parbox[t]{20mm}{\centering\textbf{Solutions}}} & 
			\multirow{2}*{\parbox[t]{6mm}{\centering\textbf{Ref.}}} &  
			\multirow{2}*{\parbox[c]{100mm}{\centering \textbf{Description}}}&
			\multirow{2}*{\parbox[t]{20mm}{\centering\textbf{Technologies}}} \\
			
			&&& \\

			\hline
			
			\multirow{10}*{\parbox[c][10mm]{20mm}{\centering Visual-based explanations}} & 
			{\centering ~\cite{9714736}} & {\parbox[c][10mm]{100mm}{ A specific interpretation technique designed for EEG signal detection models to reveal localized regions of input signals that are important for prediction.}}&{\parbox[c][10mm]{20mm}{\centering CAM}}\\
			
			\cline{2-4}
			&{\centering ~\cite{9349455}} 
			&{\parbox[c][10mm]{100mm}{ A novel interpretable multi-modal DL model to recognize important biomarkers in brain images.}}&{\parbox[c][10mm]{20mm}{\centering Grad-CAM}}\\
			
			\cline{2-4}
			&{\centering ~\cite{9435063}} 
			&{\parbox[c][10mm]{100mm}{An interpretable lightweight CNN design based on an attention mechanism.}}&{\parbox[c][10mm]{20mm}{\centering Attention mechanism}}\\
			
			\cline{2-4}
			&{\centering ~\cite{9870679}} 
			&{\parbox[c][10mm]{100mm}{A visual attention model generates fine-grained causal visual attention heat maps to explain the steering control of AVs.}}&{\parbox[c][10mm]{20mm}{\centering Attention mechanism}}\\
			
			
			\hline
			
			\multirow{4}*{\parbox[c][10mm]{20mm}{\centering Feature-based explanations}} & 
			{\centering ~\cite{9830113}} &  {\parbox[c][10mm]{100mm}{  A DL framework based on the SHAP approach to provide local and global interpretation for transparent and robust intrusion detection.}}&{\parbox[c][10mm]{20mm}{\centering SHAP}}\\
			
			\cline{2-4}
			&{\centering ~\cite{9464703}} 
			& {\parbox[c][10mm]{100mm}{ An interpretable framework for integrated predictive modeling based on LIME and SHAP to explain the importance of relevant features.}}&{\parbox[c][10mm]{20mm}{\centering SHAP, LIME}}\\
			
			\hline
			
			\multirow{4}*{\parbox[c][10mm]{20mm}{\centering Surrogate model-based explanations}} 
			&{\centering ~\cite{wu2018beyond}} & {\parbox[c][10mm]{100mm}{ A tree regularization technique for approximating complex decision boundaries of DL models, to provide interpretability and maintain predictive accuracy.}}&{\parbox[c][10mm]{20mm}{\centering Regularization}}\\

			\cline{2-4}
			&{\centering ~\cite{9933617}} & {\parbox[c][10mm]{100mm}{A fuzzy local alternative model to improve the interpretability of learning model results.	}}&{\parbox[c][10mm]{20mm}{\centering Fuzzy rule}}\\

			\hline
			
		\end{tabular}%
		\label{tab7:addlabel4}%
		\vspace{-3mm}
	\end{table*}
	
	
	\indent The transparency of decision-making processes allows users to trust the model, while allowing system administrators to monitor and adjust the model behavior. According to the description in subsection \uppercase\expandafter{\romannumeral3}-C-3, although deep models used for edge training and inference have undergone compression and processing, it is still a complex black-box model. Therefore, achieving intrinsic interpretability of models becomes extremely difficult. In contrast to model intrinsic interpretability, model decision interpretability focuses on explaining specific decisions without attention to the interpretability of the entire model structure and parameters. In subsection \uppercase\expandafter{\romannumeral3}-B, we introduce several widely used interpretation techniques, which can be broadly classified into three categories: vision-based interpretation, feature-based interpretation, and alternative model-based interpretation. These approaches focus on revealing key factors that lead to particular outputs of the model, thereby improving the understanding and interpretation of model decisions. As shown in Tab. \ref{tab7:addlabel4}, we briefly summarize the literature related to interpretable solutions of decision-making.
	
	\indent \textbf{Visual-based explanations} offer an intuitive and visual representation of decision-making processes and underlying patterns of models. For instance, authors in~\cite{9714736} refine the CAM technique to highlight relevant portions of EEG signals associated with mental states in the vehicle drowsiness monitoring system. Authors in~\cite{9349455} utilize the guided-Grad-CAM method to provide real-time explanations with high-resolution activation maps for multi-modal DL models. Different from CAM, authors in~\cite{9435063} utilize attention modules for fine-grained spatial localization, generating accurate heatmaps. Furthermore, research in~\cite{9870679} combines micro and macro interpretation modules to explain the failure cases of object detection models in autonomous driving systems. By extracting and visualizing features of CNNs and providing spatio-temporal information, this method aims to assist model developers in understanding, fine-tuning, and developing models.

	\indent\textbf{Feature attribution} methods provide explanations by analyzing the importance of features that affect the output. LIME and SHAP are two commonly used feature-based post-hoc interpretation methods. Authors in~\cite{9830113} use DeepSHAP method to provide local and global explanations for DL-based intrusion detection systems in IoV networks. Differently, authors in~\cite{9464703} combine LIME and SHAP to explain the importance of clinical features and genotype in warfarin daily dose prediction models. This combination also performs global and local interpretations to help healthcare practitioners understand and trust model predictions. In the global interpretation, the ranked importance and SHAP interpreter produce a ranking of feature importance across the entire dataset. In local interpretation, LIME and SHAP interpreters show the effect of features on the output of models run on specific samples.

	\textbf{Surrogate models} approximate the behavior of black box models and provide explanations. Authors in~\cite{wu2018beyond} propose a tree regularization technique to approximate the complex decision boundaries of deep models. By optimizing the surrogate models, domain experts can gain a good understanding of black box models' behavior. In contrast, authors in~\cite{9933617} propose a fuzzy rule-based local agent model that is capable of providing a model-independent interpretation. In addition, the fuzzy rule-based model allows one to strike a balance between prediction accuracy and the number of rules. Once established, the granularity of information in the fuzzy rule backparts can be interpreted, transforming lengthy sequences of numbers into concise and detailed descriptors, and endowing the generated fuzzy rules with a high degree of interpretability.
	

	
	
	\indent \textbf{Lesson 3:} In this subsection, we summarize interpretable implementations of lightweight models deployed at the network edge to resolve the complex black-box processing problem. This not only enhances transparency in EI systems, but also ensures that the decisions are rational and ethically responsible. In terms of decision interpretability, vision-based interpretation may be more intuitive for certain tasks, while feature-based interpretation is more suitable for scenarios that require detailed characterization. Multiple interpretation methods can also be combined with each other to provide users with a comprehensive interpretation of the model. It is important to note that ensuring the security of interpretation techniques is also an important direction for future research. Attackers can utilize interpretable techniques to detect vulnerabilities, especially in risk-sensitive scenarios~\cite{slack2020fooling}. In addition, there is a lack of scientific evaluation systems to assess interpretation methods.
	
	
	
	\subsection{Solution for Sustainability}
	
	\indent In this subsection, we delve into solutions for the trade-off between energy consumption and QoS, and the collection and storage of high-quality data. These solutions not only address the performance imbalance caused by edge nodes with limited resources, but also help ensure their sustainable operations.
	
	\subsubsection{Trade-offs Between Energy Consumption and QoS}

	\indent Communication and computation delays directly affect the responsiveness of AI applications and is particularly important for tasks that require real-time decision making and fast feedback, such as AVs and industrial automation. By considering both latency and energy consumption, systems can minimize resource consumption while delivering high-quality services, thus achieving sustainability in`edge environments. As shown in Tab.~\ref{tab8:summary}, we provide a brief overview of solutions on sustainability.
	
	\indent Authors in~\cite{10129089} consider both proximity constraints and capacity constraints of edge servers, aiming to maximize the number of users served while minimizing system energy consumption. They propose an energy saving policy that is dynamically formulated over time, deciding the state in which the edge server is woken up or put into hibernation. A heterogeneous framework is proposed in~\cite{zhu2021green}, which organizes various heterogeneous resources and is designed to achieve efficient utilization and intelligent scheduling, as well as provide a reliable and sustainable environment. The node scheduler is responsible for recording the status and information of all edge nodes, including the current workload, hardware specifications, and software settings. When an AI task request is received, the system find the most suitable edge node for execution to minimize the processing energy consumption while meeting the latency requirements. 
	
	\begin{table*}[tbp]
		\caption{Balanced energy consumption and QoS solutions for sustainability.}
		\label{routing}
		\centering
		\begin{tabular}{|c|p{75mm}|p{30mm}|p{30mm}|} 
			\hline

			\multirow{2}*{\parbox[c]{5mm}{\centering\textbf{Ref.}}} &
			\multirow{2}*{\parbox[c]{75mm}{\centering \textbf{Description}}} &
			\multirow{2}*{\parbox[c]{30mm}{\centering \textbf{Network scenarios}}} &
			\multirow{2}*{\parbox[c]{30mm}{\centering \textbf{Optimization Metrics}}}
			
			\\

			\hline
			

			{\centering~\cite{10129089} } &

			{\parbox[c][10mm]{75mm}{A dynamic scheduling strategy for energy-efficient MEC services.}}&
			
			
			\parbox[c][10mm]{30mm} {\centering MEC networks} &
			\parbox[c][10mm]{30mm} {\centering Energy consumption and user coverage}
			\\
			\hline
			
			{\centering~~\cite{zhu2021green}	 } &

			{\parbox[c][10mm]{75mm}{An online scheduling strategy for energy-efficient AI services.}}&
			
			
			\parbox[c][10mm]{30mm} {\centering  IoT } &
			\parbox[c][10mm]{30mm} {\centering Energy consumption and latency}
			\\
			\hline
			
			{\centering ~\cite{9475471}} &

			{\parbox[c][10mm]{75mm}{An efficient resource allocation and fair task offloading scheme for sustainable services in UAV-assisted MEC networks.
			}}&
			
			
			\parbox[c][10mm]{30mm} {\centering UAV-MEC networks} &
			\parbox[c][10mm]{30mm} {\centering Energy consumption and latency}
			\\
			\hline
			
			{\centering ~\cite{9968236}} &

			{\parbox[c][10mm]{75mm}{A real-time resource allocation scheme based on DRL algorithm to reduce energy consumption and system latency in UAV networks.
					
			}}&
			
			
			\parbox[c][10mm]{30mm} {\centering UAV-MEC networks} &
			\parbox[c][10mm]{30mm} {\centering Latency, energy consumption and service coverage}
			\\
			\hline
			{\centering ~\cite{9984691}} &

			{\parbox[c][10mm]{75mm}{An energy efficient edge collaborative AI model inference framework based on multi-agent RL.
			}}&
			
			
			\parbox[c][10mm]{30mm} {\centering Edge inference networks} &
			\parbox[c][10mm]{30mm} {\centering Energy consumption, latency, and accuracy	
			}
			\\
			\hline
			
			{\centering \cite{9796588}} &

			{\parbox[c][10mm]{75mm}{A lightweight feature compression method for fast and energy-efficient collaborative inference.
			}}&
			
			
			\parbox[c][10mm]{30mm} {\centering Edge inference networks} &
			\parbox[c][10mm]{30mm} {\centering Latency and energy consumption
			}
			\\
			\hline
			{\centering ~\cite{10250981}} &

			{\parbox[c][10mm]{75mm}{A collaborative inference framework for transformer model decomposition based on knowledge distillation.
			}}&
			
			
			\parbox[c][10mm]{30mm} {\centering Edge inference networks} &
			\parbox[c][10mm]{30mm} {\centering Energy consumption, latency, and accuracy
			}
			\\
			\hline
			
			{\centering ~\cite{10301516}} &

			{\parbox[c][10mm]{75mm}{A comprehensive framework for implementing edge deployment based on the transformer model.}}&
			
			
			\parbox[c][10mm]{30mm} {\centering Edge inference networks} &
			\parbox[c][10mm]{30mm} {\centering Energy consumption, latency, and accuracy
			}
			\\
			\hline
			
			{\centering ~\cite{9916128}} &

			{\parbox[c][10mm]{75mm}{A flexible weight quantification approach for energy-efficient FL.}}&
			
			
			\parbox[c][10mm]{30mm} {\centering Edge training networks} &
			\parbox[c][10mm]{30mm} {\centering Energy consumption, latency, and accuracy
			}
			\\
			\hline
			
			{\centering ~\cite{9461628}} &

			{\parbox[c][10mm]{75mm}{An energy-efficient FL based on CPU-GPU heterogeneous computing.}}&
			
			
			\parbox[c][10mm]{30mm} {\centering Edge training networks} &
			\parbox[c][10mm]{30mm} {\centering Energy consumption, latency, and accuracy
			}
			\\
			\hline
			
			{\centering~\cite{xu2023energy}} &

			{\parbox[c][10mm]{75mm}{A novel framework to minimize energy consumption of FL in MEC networks, where the availability of UEs is uncertain.}}&
			
			
			\parbox[c][10mm]{30mm} {\centering Edge training networks} &
			\parbox[c][10mm]{30mm} {\centering Energy consumption and accuracy
			}
			\\
			\hline
			
			{\centering~\cite{9605599}} &

			{\parbox[c][10mm]{75mm}{A dynamic device scheduling algorithm with energy-aware FL frameworks.}}&
			
			
			\parbox[c][10mm]{30mm} {\centering Edge training networks} &
			\parbox[c][10mm]{30mm} {\centering Energy consumption and accuracy
			}
			\\
			\hline

		\end{tabular}
		\label{tab8:summary}%
		\vspace{-5mm}
	\end{table*}
	
	\indent In many scenarios, fixed edge servers are often overloaded in hotspots and expensive to deploy in remote areas, making the above approaches difficult to realize in specific contexts. Therefore, UAV-based EI architecture becomes a viable solution, especially for areas where service resources are scarce. For example, flexible UAVs equipped with AIGC servers enable users to access AIGC services with ultra-low latency and high reliability~\cite{xu2023unleashing}. Therefore, the energy-efficient design of UAVs is important to ensure their sustainable operation. Authors in~\cite{9475471} focus on sustainable services in UAV-assisted MEC networks. This article aims to minimize energy consumption of UAVs by jointly optimizing flight trajectories, resource allocation, and task scheduling with fairness. Differently, authors in~\cite{9968236} not only consider energy consumption of UAVs, but also focus on uplink transmission of mobile users. They use DRL algorithms to solve the joint optimization problem of UAV motion control, mobile user association and power control.
	
	\indent In deploying large model services to the network edge, the above resource scheduling and offloading strategies can be utilized to balance the overall service latency and device energy consumption. Furthermore, some studies focus on joint optimization of latency, accuracy and energy consumption of model training and inference processes. For instance, authors in~\cite{9984691} present a multi-agent RL collaborative inference scheme that enables each device to select the best DNN division point and collaborative edge based on the number of images, channel conditions and previous inference performance. The Q-values of neighboring devices are used to accelerate the optimization of the inference strategy through learning experience exchange, thus reducing energy consumption while ensuring inference latency and quality. Differently, authors in~\cite{9796588} propose auto-encoder based intermediate feature compression to reduce the communication overhead in collaborative inference  thereby achieving fast and energy-efficient edge inference. 
	
	\indent Unlike~\cite{9984691} and~\cite{9796588}, studies in~\cite{10250981} and \cite{10301516} focus on transformer models in the edge inference process. These studies are important for efficient and accurate reasoning based on transformer models, such as mobile AIGC. Authors in~\cite{10250981} decompose large vision transformers into multiple smaller models to be deployed at edge devices to reduce inference latency and energy consumption. In addition, the loss of model accuracy caused by decomposition is reduced by an algorithm based on knowledge distillation. Authors in~\cite{10301516} introduce multiple frameworks to enable efficient deployment of transformer architectures to EI platforms. Specifically, ProTran and FlexiBERT 2.0 are used for modeling accuracy as well as latency, energy consumption, and hardware measurements. EdgeTran employs alternative models derived from ProTran and FlexiBERT 2.0 to obtain the best-performing model-device pairs and performs the optimization of the output model through block-level growth and castration techniques.

	
	
	\indent Moreover, as for collaborative training frameworks across edge devices such as FL, local computation and frequent communication may overwhelm energy-constrained mobile devices. A possible solution is to develop multiple model compression methods such as pruning and quantization. For example, in the study of~\cite{9916128}, an iterative algorithm for jointly determining weight quantization and spectrum resource allocation strategies among devices is proposed.
	
	\indent Some studies address the balance between learning performance and energy consumption by resources management. Authors in~\cite{9461628} achieve energy-efficient FL in wireless networks by considering the allocation of both computational and communication resources on four dimensions: bandwidth allocation, time partitioning, CPU-GPU workload partitioning, and CPU-GPU frequency scaling. Authors in~\cite{xu2023energy} leverage complex interactions among environmental contextual information (e.g., workload, the amount of available computational resources, and data quality) to select available User Equipments (UEs) and propose an approximation algorithm to find the suitable aggregator location. Similarly, authors in~\cite{9605599} also introduce an energy-aware device scheduling algorithm. Unlike~\cite{xu2023energy}, this article employs an analog gradient aggregation scheme, which aims to achieve more efficient device scheduling by aggregating local updates over the same time-frequency resource blocks.

	\subsubsection{High-quality Data Collection and Storage}
	
	\indent High-quality data can fuel the training, inference, and optimization of AI models. Therefore, data collection and sustainable data storage are important to ensure reliable network services.
	
	\indent \textbf{Data collection:} Mobile Crowd Sensing (MCS) utilizes smart devices and sensors that are widely distributed at the network edge to cellularize data to support various intelligent applications. However, it becomes difficult to directly assess the quality of collected data without any prior knowledge. To address the issue, a data quality-based MCS incentive mechanism is designed in~\cite{7946149}, which aims to enable long-term effective contributions by paying participants who provide high-quality data. Authors in~\cite{9714882} propose a cost and quality-aware data collection scheme for edge-assisted vehicle crowd-sensing systems. They employ adaptive clustering and online sensing parameter tuning to avoid unnecessary data collection while ensuring reliable and timely data uploads.
	
	\indent Authors in~\cite{9686356} utilize blockchain for data quality control. The blockchain records key information about the employment relationship, participants and consensus nodes to ensure that the data is undeniable and tamper-proof. Accurate reward payments are realized based on user data quality, truth results, reward rules and employment relationships. 
	
	\indent In addition, multi-modal data fusion helps to integrate information from different modalities, including vision, speech, and sensor data. Authors in \cite{10160040} combine modal data fusion techniques and GAN to build a cross-modal data generator. The generator can generate long-term time series data from spatial-temporal modal data, and then replace missing values with the generated data.
	
	\indent \textbf{Data storage:} A reliable and scalable storage solution ensures data persistence and availability. Authors in~\cite{8654013} focus on ensuring cache fairness among peers in edge environments. They propose approximation algorithms and distributed algorithms to determine cache nodes. Further, fairness metrics are utilized to achieve continuous caching decisions over time. Authors in~\cite{9488804} explore various criteria, such as data prevalence and proximity to edge processing functions, to strategically allocate diverse classes of raw and processed data at the network edge. They provide a lightweight mechanism to identify data types, data generation sources, and supported processing functions.
	
	\indent Due to the limited storage capacity of edge servers, reducing data redundancy is of great importance to improve the storage utilization of edge servers~\cite{10049532}. Deduplication methods are widely used to reduce data redundancy in storage systems. Authors in~\cite{10049532} use integer programming and Lagrangian relaxation methods, as well as an improved subgradient approach to solve the balanced deduplication problem. In the work presented in~\cite{10219047}, the simultaneous consideration of file popularity, file similarity, and server reliability aims to enhance the availability of popular files while minimizing unnecessary space redundancy. The article underscores the adverse effects of server unreliability on storage hit rates and addresses this issue by implementing similarity-aware hierarchical clustering algorithms.

	\indent \textbf{Lessons 4:} In this subsection, we focus on two key perspectives, efficient energy utilization and data quality, to facilitate sustainable and long-term service quality in resource-constrained edge environments. This is crucial for the efficient deployment and inference of AI models at the network edge. By recognizing that optimizing one factor may lead to sacrifices in other aspects, striking a balance between energy consumption and service quality empowers the system to provide high-quality services while minimizing resource consumption, thereby enhancing the overall sustainability of the EI system. 
	
	\indent  Additionally, due to the varying quality of data, including noise and biases, there can be impacts on the accuracy of large models. Therefore, multi-modal data fusion and GANs can provide rich and realistic data representations for various tasks. This approach is also helpful in learning complex relationships among data to support advanced applications. Nevertheless, research on these technologies is still in its infancy.
	
	\section{Research Challenges and Open Issues}\label{Research Challenges and Open Issues}
	
	\indent Research on trustworthy EI is still in its infancy. In order to achieve the vision of secure, reliable, transparent, and sustainable trustworthy EI, there are many issues and research directions that need to be further explored. Inspired by existing solutions, in this section, we discuss some research challenges and open issues in the field of trustworthy EI.
	
	\subsection{Security Enhancement}

	\subsubsection{Security and Privacy in D2D-based Decentralized Paradigm}
	
	\indent Decentralized EI relies solely on communication and consensus among end devices to process tasks, thus overcoming the dependence on central nodes while reducing the high cost of communications with edge servers. However, in this scenario, there are several security and privacy risks, such as: \textit{\romannumeral1}) Decentralized entity control: Different nodes in a decentralized system may be controlled by distinct administrative entities, leading to potential inconsistencies in the security level of nodes. A vulnerable node may become a target of attackers, posing a threat to the overall system security; \textit{\romannumeral2}) Widespread semi-honest nodes: They can introduce instability into the system, and attempt to steal data or send malicious data during collaborative learning to disrupt system availability; \textit{\romannumeral3}) Topology diversity: The topology of a distributed system can be highly complex, with diverse relationships among nodes, complicating security analysis and management. Therefore, robust decentralized consensus algorithms need to be developed as well as optimized topology management strategies for distributed systems, while constrained resources of devices need to be considered.
	
	\subsubsection{Deployment of Zero-Trust Security at Network Edge}
	
	\indent Incorporating the zero-trust principle in EI architectures is an effective way to address both known and unknown threats. However, due to limited edge resources, it faces the following challenges and open issues: \textit{\romannumeral1}) Computational and storage resource requirements: zero-trust model requires continuous monitoring and verification, imposing an excessive burden on the limited computational and storage resources of edge devices, and a lightweight implementation needs to be found; \textit{\romannumeral2}) Low-latency requirements: EI services typically require low latency, but multiple validation steps introduced by the zero-trust model increase latency; \textit{\romannumeral3}) Diversity of edge devices: Considering the diversity of devices in edge environments, including different hardware and software specifications, the realization of zero-trust needs to be deployed consistently and efficiently on various types of devices.
	
	\subsubsection{Generative Model-based Security}
	
	\indent In the field of EI, generative models such as GANs and Variational AutoEncoder (VAE) are widely used to improve the security of systems. These generative models are capable of capturing unusual patterns or behaviors and detecting inconsistencies with normal behavior by learning normal data distributions, so that potential security threats can be detected in a timely manner. In adversarial defense of images, these methods have achieved remarkable results. However, there are challenges in migrating these approaches to communication network defense for EI. Communications in EI systems involve a wider range of data types and complex communication patterns, thus how to effectively apply generative models to detect anomalous behaviors is an issue that requires in-depth research. On the other hand, by using intelligent game-adversarial techniques, it is possible to realize the induced detection computation of EI models by "generative robots". By automating the generation of large test sets, these "generative robots" are able to detect model weaknesses and security issues. This approach provides a cutting-edge research direction for discovering and providing feedback on security issues in EI systems.
	

	\subsection{Autonomous Collaboration for Edge Co-inference}
	
	\indent Autonomous collaboration among edge nodes is essential for dynamic edge co-inference, but its implementation poses challenges in distributed training and execution. In distributed control scenarios, edge nodes can make independent decisions and execute tasks in a dynamic environment while collaborating to achieve common goals. To realize such autonomous decision-making and distributed collaboration, current research typically uses distributed RL. 
	
	\indent Fully distributed RL is particularly challenging, because it needs to consider not only the interaction between individual agents and the environment, but also the interplay among multiple agents. Correspondingly, various challenges are introduced, including: \textit{\romannumeral1}) Changes in agent strategies: It can lead to environmental instability, since the behavior of one agent affects that of others; \textit{\romannumeral2}) Distributed training and reward feedback: Distributed training requires individual agents to receive separate reward feedback. Decomposing feedback from the environment into rewards for each agent and quantifying contributions of each agent to teamwork are rather complex; \textit{\romannumeral3}) Curse of dimensionality: When the number of agents increases, the learning process faces challengs such as the curse of dimensionality, resulting in a significant increase in computational complexity. 
	
	\subsection{Large Models in Trustworthy EI}
		\indent Large models, especially LLMs, are deployed at the network edge as model collaboration control center of EI system, which not only coordinates multiple intelligent applications, but also ensures their consistency, accuracy, and real-time performance in handling complex tasks. This deployment helps to build trustworthy EI system and provide users with secure and reliable intelligent services. Specifically, EI systems will integrate multiple intelligent applications, such as AI assistants that are not only limited to smart home management, but also include real-time environment monitoring, user communication, and so on. These complex tasks are beyond the solution capability of a single AI model. Smart speakers, home control screens, smart phones, and other IoT devices are expected to become the interaction portal of Jarvis style smart housekeeper. Therefore, considering that immediacy, reliability, privacy, and computational capability, it is crucial to deploy large models as a collaborative control center for models on edge devices. However, this work is still in its infancy, and further breakthroughs are needed in compression techniques to achieve efficient deployment and support tens of billions of parametric model inference at the network edge. Key challenges include low-latency inference that matches endpoint computing capability, the impact of accuracy loss on user QoS, and the sustainability challenge of energy consumption.

	\subsection{Trade-off Between Interpretability and Accuracy at The Network Edge}
	
\begin{figure}[tbp]
		\begin{center}
			\includegraphics[scale = 0.48]{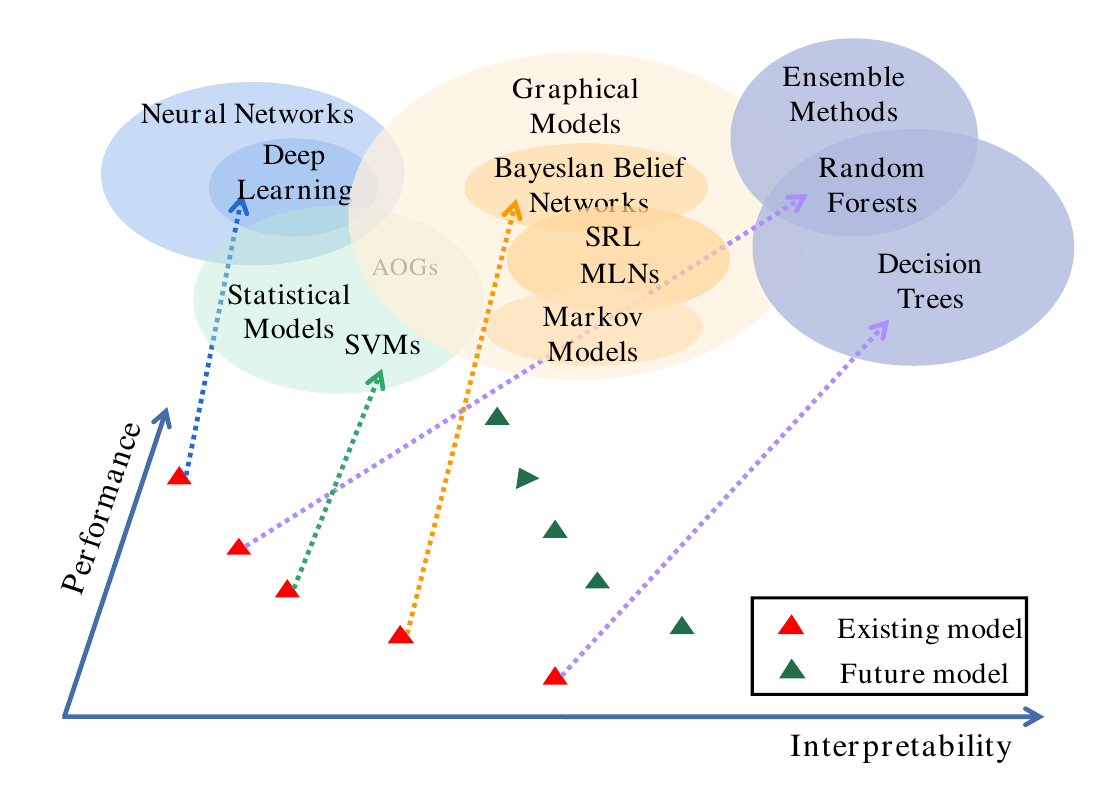}
			\caption{\centering Accuracy of interpretable models~\cite{gunning2019xai}.}\label{5:Structure of the performance}
			\vspace{-5mm}
		\end{center}
	\end{figure}

	\indent In EI, balancing accuracy and interpretability is an important issue~\cite{nussberger2022public}. Some models can achieve high accuracy, but it is difficult to explain their internal mechanisms, limiting their applications in various fields. Other models with high interpretability may sacrifice accuracy, which also affects their effectiveness in practical applications. Fig. \ref{5:Structure of the performance} shows the relationship between accuracy and interpretability of ML models.
	
	\indent This trade-off issue involves several challenging factors: \textit{\romannumeral1}) "Interpretability" is difficult to be defined and measured because different application fields and scenarios have distinct requirements; \textit{\romannumeral2}) In EI application fields like healthcare, protecting data privacy is crucial, limiting the use of models with strong interpretability; \textit{\romannumeral3}) Some models with strong interpretability may be too simplistic to handle large-scale and high-dimensional data, potentially sacrificing accuracy; \textit{\romannumeral4}) Some models may require many parameters or complex structures to achieve high accuracy, which are hard for interpretation; \textit{\romannumeral5}) In scenarios involving high-dimensional and complex data, even models with strong interpretability may struggle to explain their internal mechanisms and decision-making processes, necessitating the development of novel interpretative methods. In conclusion, achieving a balance between accuracy and interpretability in trustworthy EI is a complex task, requiring the considerations of multiple factors and practical application requirements. 
	
	\subsection{Data Governance Issues in Trustworthy EI}
	
	\indent EI integrates sensing, communication, and computation, which corresponds to the collection, transmission, and utilization of data. High-quality data resources are crucial for models to accurately understand and respond to various scenarios, especially generative AI large models that require a large amount of high-quality data to enhance content accuracy. However, multi-modal data contains rich societal knowledge. For example, digital data, as a potentially huge resource, is difficult to apply for big model training due to its simple expression form and lack of linguistic features. In addition, utilizing generated data to train AI models will lead to model degradation. Specifically, the generated data obtained by sampling the output of the generative model will lose some information. This situation can gradually accumulate and ultimately lead to a distribution of the generated data that bears little resemblance to the real data. As a result, models trained with such generated data will lead to degraded model performance and unreliable inference results~\cite{shumailov2023curse}. Therefore, the efficient fusion of multi-modal data and high-quality synthetic data will become the difficulty of breakthrough.
	
	\indent Furthermore, ensuring visibility into data use and sharing can help to improve users' understanding of how data are used and increase their sense of control and trust in data privacy. However, achieving such visibility requires strong technical support, including real-time monitoring, logging and permission tracking. Overcoming these technical challenges requires significant investment in resources and technology development. Finding the suitable balance between improving visibility while balancing the user's right to privacy and the protection of sensitive information, such as trade secrets, is a complex task.
	
	\section{Conclusion}\label{Conclusion}
	
	\indent We summarize and discuss the development, solutions and challenges in a large number of related literatures on trustworthy EI. First, we discuss the definition, characteristics, and architecture of trustworthy EI. We also sort out four important issues in the implementation of trustworthy EI and summarize the key enabled techniques to achieve the trustworthiness of EI. Subsequently, we conduct a comprehensive investigation from different aspects, i.e., balanced security and privacy protection, reliability, transparency, and sustainability of trustworthy EI. Finally, we discuss the relevant research challenges and open issues toward achieving trustworthy EI. We hope this survey provides an effective guideline that can inspire researchers to advance trustworthy EI.

	\bibliographystyle{IEEEtran}
	\bibliography{ref1}
\end{document}